\documentclass{article} 

\usepackage{iclr2027_conference,times}

\usepackage{wrapfig}
\usepackage{booktabs}
\usepackage{caption}
\usepackage{tabularray}
\UseTblrLibrary{booktabs}
\usepackage{url}
\usepackage{multirow}
\usepackage{graphicx}
\usepackage[table]{xcolor}
\usepackage{float}
\usepackage{enumitem}
\usepackage{tabularx}
\usepackage{amsmath,amssymb}
\usepackage[most]{tcolorbox}
\usepackage{siunitx}

\newtcolorbox{promptbox}[1][]{
  enhanced,
  breakable,
  colback=gray!5,
  colframe=black!40,
  boxrule=0.6pt,
  arc=2pt,
  left=6pt,
  right=6pt,
  top=6pt,
  bottom=6pt,
  #1
}

\usepackage{hyperref}
\hypersetup{
  colorlinks=true,
  linkcolor=blue,
  filecolor=magenta,
  urlcolor=cyan,
}
\title{SyncReward: Human-Aligned Synchronization Reward for Audio-Visual Generation}

\author{
\textbf{Kaisi Guan}\textsuperscript{1,2}\thanks{The work was done during an internship at Tencent.} \quad
\textbf{Ying Ba}\textsuperscript{1} \quad
\textbf{Wenjing Wang}\textsuperscript{2} \quad
\textbf{Xihua Wang}\textsuperscript{2} \quad
\textbf{Chen Li}\textsuperscript{2} \\
\textbf{Jing Lyu}\textsuperscript{2} \quad
\textbf{Ruihua Song}\textsuperscript{1}\thanks{Corresponding Author.} \quad
\\
\normalfont
\textsuperscript{1}\,Gaoling School of Artificial Intelligence,
Renmin University of China \\
\textsuperscript{2}\ WeChat, Tencent Inc. \\
\texttt{guankaisi@ruc.edu.cn}
}

\begin{document}

\iclrfinalcopy

\maketitle
\thispagestyle{plain}
\pagestyle{plain}

\begin{figure}[h]
  \centering
  \includegraphics[width=\textwidth]{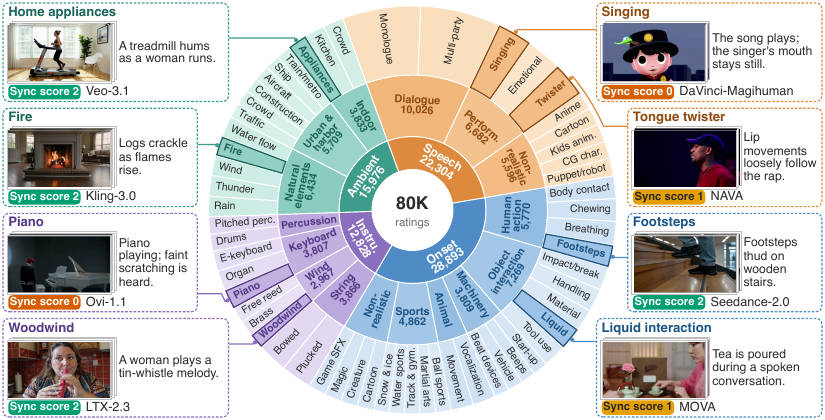}
  \caption{\textbf{Illustration of SyncReward-Data-80K}: human annotations of synchronization in generated audio--visual clips. The dataset contains 80K professional synchronization ratings on outputs from eight generators, covering four sound-event families: Speech, Onset, Instrument, and Ambient. A three-level taxonomy organizes these events into 16 classes and 58 fine-grained categories. Surrounding examples show sampled frames and brief descriptions of visible sound sources and their associated sounds, linked to the corresponding categories.}
  \label{fig:teaser}
\end{figure}
\begin{abstract}

Recent years have witnessed rapid progress in joint audio--visual generation. However, audio--visual synchronization remains a challenge, as generated speech and sound often temporally misalign with visual events. A key bottleneck is the lack of human-annotated data, reliable evaluation benchmarks, and synchronization-aware evaluation models for generated audio--visual outputs. To address these limitations, we conduct a comprehensive study spanning data, benchmarking, and reward modeling. We first introduce \textbf{SyncReward-Data-80K}, a dataset containing 80K professional synchronization ratings for clips from eight audio--visual generators. A taxonomy-guided construction pipeline covers four sound-event families: Speech, Onset, Instrument, and Ambient. We further construct \textbf{SyncReward-Bench}, a held-out benchmark of 1K clips, each independently rated by multiple professionals. Using this benchmark, we systematically evaluate specialized synchronization metrics and omni-modal models, revealing their substantial misalignment with human judgments. Finally, we develop \textbf{SyncReward}, a reward model that combines semantic--temporal contrastive pretraining on real clips with human supervision on generated clips. As an evaluation metric, SyncReward achieves the strongest agreement with human ratings among the evaluated methods. As a reward, it improves synchronization through inference-time best-of-$N$ selection and reinforcement learning based post-training. We will release the data, benchmark, model, and code at \url{https://github.com/WeChatCV/SyncReward}.
\end{abstract}

\section{Introduction}
\label{sec:introduction}


Recent advances in joint audio--visual generation~\citep{low2025ovi,hacohen2026ltx,openmoss_mova_2026,minimaxh3} have enabled models to produce realistic videos together with speech, music, and sound effects. However, even state-of-the-art models~\citep{seedance,kling,veo} still exhibit noticeable audio--visual synchronization failures. For example, a speaker's voice may continue while the mouth remains still, or an impact sound may occur before the corresponding object contact. These errors make generated videos less realistic. Detecting them is important for evaluating and improving audio--visual generation.

Existing approaches, however, remain limited for this purpose. Specialized synchronization metrics~\citep{yariv2024avalign,iashin2024synchformer,luo2023cavp,guan2026fateframelevelaudiovisualtemporal} estimate temporal offsets, or build audio and visual representations. While useful, these proxy objectives may not fully reflect human judgments on generated audio--visual content, where synchronization failures can be semantic ambiguity, missing or hallucinated events rather than simple global temporal shifts. General-purpose omni-modal large language models~\citep{Qwen2.5-Omni,xu2025qwen3omnitechnicalreport} offer strong semantic understanding, but are not  trained to capture fine-grained temporal misalignments.

To address these limitations, we provide a comprehensive study covering data construction, benchmarking, and reward model design.
First, we introduce \textbf{SyncReward-Data-80K}, a large-scale human-annotated dataset for studying synchronization in generated audio--visual content.
A three-level sound-event taxonomy guides caption sampling and prompt refinement, covering Speech, Onset, Instrument, and Ambient events. We then collect outputs from eight proprietary and open-source audio--visual generation models, capturing diverse generation behaviors and failure patterns. Finally, professional annotators evaluate audio--visual synchronization with a three-level rating protocol, yielding human synchronization annotations for 80K generated clips. Overall, the dataset provides systematic semantic coverage, diverse model outputs, and reliable human supervision.



We further introduce \textbf{SyncReward-Bench}, a disjoint evaluation set of 1,000 generated clips across sound-event categories and generators, each independently rated by multiple professional annotators. Using this benchmark, we comprehensively evaluate synchronization metrics, off-the-shelf omni-modal judges, and fine-tuned omni-modal models. Extensive experiments show that existing evaluators remain limited, while fine-tuning on human-rated generated clips substantially improves agreement with human judgments, highlighting the value of task-specific annotations.

Building on \texttt{SyncReward-Data-80K}, we develop \textbf{SyncReward}, a reward model for both synchronization evaluation and reward-guided generation. SyncReward first learns general temporal correspondence from real audio--visual clips via semantic--temporal contrastive learning, and is then trained to predict human synchronization ratings on generated clips. \textbf{As an evaluation metric}, it achieves the strongest agreement with human ratings across all four sound-event families on \texttt{SyncReward-Bench}, outperforming larger omni-modal models fine-tuned on the same data. \textbf{As a reward}, it guides inference-time Best-of-$N$ selection and reinforcement-learning-based post-training, yielding better synchronization than the evaluated alternative rewards in both settings.
These results demonstrate that a human-aligned synchronization reward can serve as both a reliable evaluator and an effective optimization signal for audio--visual generation.
 
Our contributions are threefold:
\begin{itemize}[leftmargin=2em]

\item \textbf{Balanced, diverse, and high-quality data.} We introduce SyncReward-Data-80K, with 80K professional annotations spanning eight audio--visual generators and four sound-event families.

\item \textbf{Comprehensive benchmarking.} We build SyncReward-Bench and systematically benchmark specialized synchronization metrics, off-the-shelf omni-modal judges, and fine-tuned models.

\item \textbf{Reliable reward modeling.} We develop SyncReward, which better aligns with human judgments and scales effectively to inference-time selection and generative post-training.

\end{itemize}
\section{Related Work}
\paragraph{Audio-Visual Joint Generation.}
Existing joint audio--visual generators can be classified into two categories. Dual-tower models~\citep{low2025ovi,hacohen2026ltx,guan2026bridgedit,openmoss_mova_2026,ji2026nava} connect modality-specific branches through cross-modal layers, allowing pretrained components to be reused. Single-tower models~\citep{davinci-magihuman-2026,magi2,minimaxh3} process audio and video tokens within a shared Transformer backbone. These architectures learn cross-modal dependencies, but their generative objectives do not directly optimize human judgments of synchronization. In this paper, we explore how to evaluate and improve audio--video synchronization without changing the generator architecture.

\paragraph{Reward Models for Generative Models.}
Reward models learn human preferences over generated outputs and provide objectives for optimization.  ImageReward~\citep{xu2023imagereward} and PickScore~\citep{kirstain2023pickapic} establish this approach for text-to-image generation. Later work broadens the assessment beyond a single preference score: VisionReward~\citep{xu2024visionreward} models multiple dimensions of image and video quality, while EditReward~\citep{wu2026editreward} targets instruction-guided image editing. VideoScore~\citep{he2024videoscore} and VideoAlign~\citep{liu2025videoreward} learn human assessments of generated videos, including visual quality, temporal consistency, and text alignment. These visual rewards do not directly assess whether sounds coincide with visible actions. In this work, we address this gap through a comprehensive study of audio--visual synchronization, including data construction, benchmarking, and reward model design.

\paragraph{Audio-Visual Synchronization Evaluation.}
Existing methods assess audio--visual synchronization from different perspectives. AV-Align~\citep{yariv2024avalign} compares audio onsets with visual motion peaks, whereas SparseSelector~\citep{iashin2022sparseselector} and Synchformer~\citep{iashin2024synchformer} predict temporal offsets. Embedding-based methods assess cross-modal similarity, including SyncNet~\citep{chung2017syncnet} for lip synchronization and CAVP~\citep{luo2023cavp} and FATE~\citep{guan2026fateframelevelaudiovisualtemporal} for broader audio--visual events. PEAVS~\citep{goncalves2024peavs} learns synchronization scores from human ratings.
Beyond Time Shifts~\citep{qian2026timeshiftsadaptingomnillm} further adapts an omni-modal LLM using SynthSync, which pairs real videos with generated audio and human preferences.
Despite these developments, existing works do not fully capture the native patterns of joint audio--video generators. We address this limitation with diverse generated outputs and high-quality human annotations, and further evaluate whether the learned score can improve generation.

\section{SyncReward}
\label{sec:method}
\begin{figure*}[t]
    \centering
    \includegraphics[width=\textwidth]{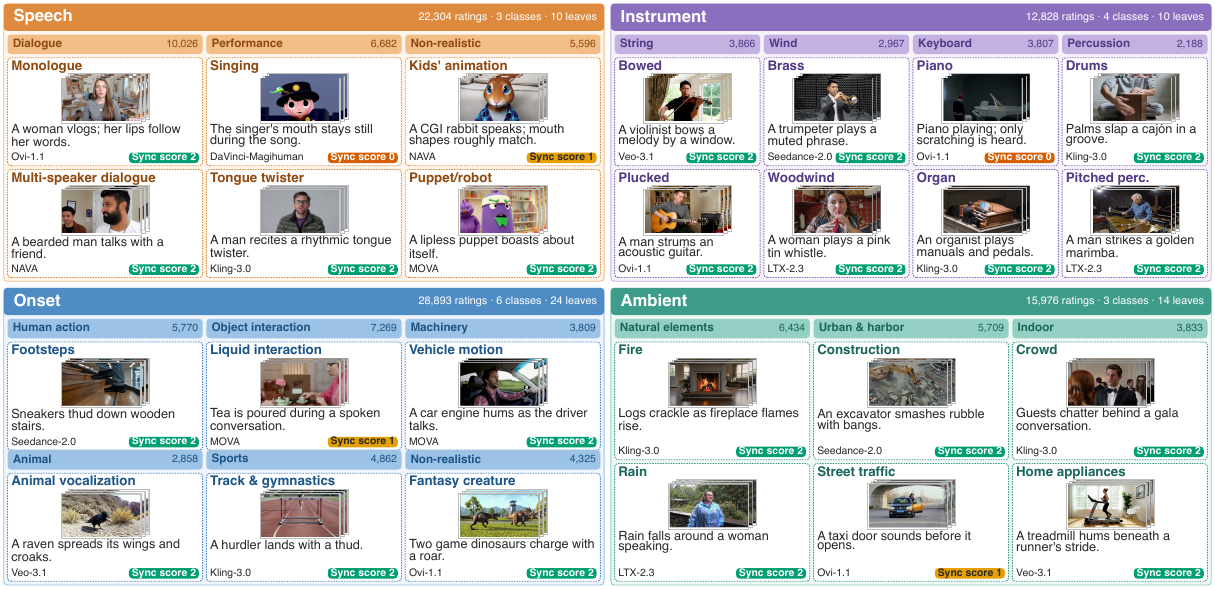}
   \caption{\textbf{Sound-event taxonomy and human-rated examples.} Examples span four sound-event families, 16 classes, and eight audio--visual generators. Headers report dataset-level rating counts for each family and class. Each card presents three sampled video frames, an event description, the generator name, and a human synchronization score. Scores indicate unsynchronized (0), partially synchronized (1), and fully synchronized (2) clips. Examples are selected to illustrate category coverage and different synchronization levels, rather than the dataset's score distribution.}
    \label{fig:representative_examples}
    \vspace{-5mm}
\end{figure*}






We first present the task definition, then detail the data collection, construction, and annotation procedures, and finally introduce the design of the reward model.

\paragraph{Task Formulation.}
Given a condition $c$, such as a text prompt, a joint audio--visual generator produces a video clip and its accompanying audio, $x=(x^v,x^a)$. We evaluate whether sounds are aligned with their visible sources, such as speech with lip motion and impact sounds with object contact. Our evaluation focuses on clips whose audio is related to the visible scene and distinguishes synchronization from perceptual quality and prompt adherence. 
Our goal is to obtain a clip-level score $R(x^v,x^a)\in[0,2]$ that agrees with human judgments of synchronization quality.
The score serves both as an evaluation metric and as a reward for candidate selection and post-training.

\subsection{SyncReward-Data-80K and SyncReward-Bench}
\label{sec:data}

\paragraph{Taxonomy.}
We use Qwen2.5-Omni-7B~\citep{Qwen2.5-Omni} to caption real audio--visual clips from AudioSet~\citep{audioset} and VGGSound~\citep{chen2020vggsound}, describing visible events and their accompanying sounds. We organize these descriptions into a three-level taxonomy comprising four event families, 16 intermediate categories, and 58 fine-grained subcategories. The four families are Speech, covering visible speaking and singing; Onset, covering sounds associated with visible actions and discrete events; Instrument, covering visible musical performances; and Ambient, covering environmental sounds with identifiable visual sources. Off-screen narration and background music are excluded from the target events.
This taxonomy guides prompt sampling and refinement and defines the categories used for evaluation.
The complete hierarchy and six additional language tags for Speech are provided in the Appendix.

\paragraph{Prompt Construction.}
The caption pool collected from AudioSet~\citep{audioset} and VGGSound~\citep{chen2020vggsound} is highly imbalanced across sound-event categories. We therefore map each caption to the taxonomy and resample the pool according to predefined category quotas, balancing category coverage with relevance to synchronization evaluation.
This produces three disjoint sets of $10$K captions and reduces the dominance of frequent categories in the original pool. We then use an LLM to rewrite each sampled caption into a generation prompt that explicitly describes the temporal correspondence between a visible sound source, its action, and the associated sound. The rewriting preserves the original scene where possible and removes sound descriptions that cannot be linked to a visible source. 
These prompts specify the intended synchronization, 
while the generated clips may still contain synchronization errors. The resulting category distribution is shown in Figure~\ref{fig:teaser};
the sampling quotas and rewriting instructions are provided in the Appendix.

\paragraph{Audio--Visual Generation.}
We synthesize audio--visual clips using eight close-source and open-source generators: Kling~3.0~\citep{kling}, Seedance~2.0~\citep{seedance}, Veo~3~\citep{veo}, LTX-2.3~\citep{hacohen2026ltx}, daVinci~\citep{davinci-magihuman-2026}, MOVA~\citep{openmoss_mova_2026}, NAVA~\citep{ji2026nava}, and Ovi~\citep{low2025ovi}. The generation pool contains approximately $10$K clips per model. 
Using multiple generators enables the dataset to cover a diverse range of synchronization error patterns rather than artifacts specific to a single generation system.

\paragraph{Human Annotation.}
Each generated clip is independently assessed by five professional annotators under a two-stage protocol, with generator identities hidden. Annotators first determine whether the audio is related to the visible scene, then assign related clips an ordinal synchronization rating $y\in\{0,1,2\}$, indicating clearly desynchronized, partially or uncertainly synchronized, and well synchronized, respectively.
Finally, the constructed \texttt{SyncReward-Data-80K} contains $80$K professional synchronization ratings. We aggregate annotations into bias-corrected consensus targets and construct training and validation splits with no shared $(\mathrm{uid},\mathrm{model})$ pairs, where $\mathrm{uid}$ identifies the source prompt. 
Representative examples can be found in Figure~\ref{fig:representative_examples}.
More details on the annotation guidelines, annotation platform, label aggregation are reported in the Appendix.

\paragraph{SyncReward-Bench.} 
We further construct \texttt{SyncReward-Bench}, an independently annotated benchmark containing $1{,}000$ generated clips. Samples are drawn across generators according to predefined taxonomy proportions and are strictly disjoint from \texttt{SyncReward-Data-80K}. Each clip is independently rated by multiple professional annotators, followed by cross-checking and score aggregation to obtain robust human reference scores.
Spearman's rank correlation and pairwise ranking accuracy measure agreement with human ratings and their induced orderings, respectively. The Appendix details benchmark composition, annotation checks, and pair construction. Section~\ref{sec:metric} compares synchronization metrics and omni-modal judges.

\subsection{SyncReward Architecture and Two-Stage Training}
\label{sec:arch}

\begin{figure*}[t]
    \centering
    \includegraphics[width=\textwidth]{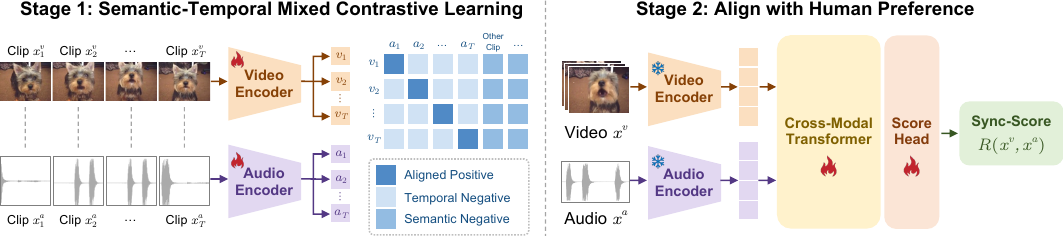}
    \caption{\textbf{Two-stage training of SyncReward.} Stage~1 trains the audio and video encoders through contrastive learning with aligned segment pairs, within-clip temporal negatives, and cross-clip semantic negatives from the same batch. Stage~2 freezes both encoders and trains a cross-modal Transformer and score head using human synchronization ratings, combining regression and pairwise ranking losses. The model outputs a clip-level synchronization score in $[0,2]$. Flame and snowflake icons indicate trainable and frozen modules, respectively.}
    \label{fig:training_pipeline}
\end{figure*}

SyncReward comprises audio and video encoders, a cross-modal Transformer, and a scalar score head.
In Stage 1, we first train the encoders to capture
semantic and
temporal correspondence using real audio--visual clips.
In Stage 2, we then freeze the encoders and train the remaining components on human-rated generated clips. The overall pipeline is illustrated in Figure~\ref{fig:training_pipeline}.

\paragraph{Stage 1: Semantic--Temporal Mixed Contrastive Learning.}
We divide each real audio--visual clip into $T$ temporally corresponding segments and encode them as $v_t=f_v(x_t^v)$ and $a_t=f_a(x_t^a)$. Video and audio segments from the same clip and temporal position form 
\emph{aligned positives}. Segments from different temporal positions within the same clip serve as \emph{temporal negatives}, while segments from different clips serve as \emph{semantic negatives}. Together, these negatives encourage the encoders to distinguish both audio--visual semantic correspondence and temporal alignment.

We further introduce a batch index $b \in \{1,2,\ldots,B\}$ for samples in a training batch, yielding $v_{b,t}$ and $a_{b,t}$.
We then optimize a symmetric contrastive objective:
\begin{equation}
\begin{aligned}
\mathcal{L}_{\mathrm{S1}}
&=
\frac{1}{2}
\left(
\mathcal{L}_{v\rightarrow a}
+
\mathcal{L}_{a\rightarrow v}
\right), \\
\mathcal{L}_{v\rightarrow a}
&=
-\frac{1}{BT}
\sum_{b=1}^{B}
\sum_{t=1}^{T}
\log
\frac{
\exp\!\left(
\operatorname{sim}(v_{b,t},a_{b,t})/\tau
\right)
}{
\displaystyle
\sum_{b'=1}^{B}
\sum_{t'=1}^{T}
\exp\!\left(
\operatorname{sim}(v_{b,t},a_{b',t'})/\tau
\right)
}.
\end{aligned}
\label{eq:temporal_contrastive}
\end{equation}
Here, $\operatorname{sim}(\cdot)$
measures embedding similarity, $\tau$ is the temperature, and $\mathcal{L}_{a\rightarrow v}$ is defined by exchanging the two modalities. Each denominator includes the aligned positive and both temporal and semantic negatives.
Within-clip negatives are particularly useful because recognizing shared semantic content alone is insufficient to distinguish temporal positions.

\paragraph{Stage 2: Align with Human Preference.}
We freeze the Stage-1 encoders and learn a cross-modal Transformer $g_\phi$ and score head $r_\psi$ using \texttt{SyncReward-Data-80K}. For a generated clip $x$, the encoders produce video and audio sequences $V=\{v_t\}_{t=1}^{T}$ and $A=\{a_t\}_{t=1}^{T}$. We concatenate these sequences with a learnable reward token $\mathrm{REW}$ and a modality separator $\mathrm{MOD}$:
\begin{equation}
\begin{gathered}
H = g_\phi\!\left([\mathrm{REW};V;\mathrm{MOD};A]\right), \\
R_{\phi,\psi}(x)
= 2\sigma\!\left(r_\psi(h_{\mathrm{REW}})\right).
\end{gathered}
\label{eq:reward_model}
\end{equation}
The output $h_{\mathrm{REW}}$ aggregates information across temporal positions and modalities. The sigmoid $\sigma(\cdot)$ maps the scalar prediction to the human rating range $[0,2]$.

We train the model with regression and pairwise ranking supervision.
For a batch of $B$ clips,
let $R_b=R_{\phi,\psi}(x_b)$ and let $y_b$ denote the aggregated human target.
We derive pairwise ranking supervision
from these targets as $\mathcal{P}=\{(i,j):1\leq i,j\leq B,\ y_i>y_j\}$. The objective is:
\begin{equation}
\begin{aligned}
\mathcal{L}_{\mathrm{S2}}
&=
\mathcal{L}_{\mathrm{reg}}
+
\lambda_{\mathrm{rank}}\mathcal{L}_{\mathrm{rank}}, \\
\mathcal{L}_{\mathrm{reg}}
&=
\frac{1}{B}
\sum_{b=1}^{B}
\operatorname{SmoothL1}(R_b,y_b), \\
\mathcal{L}_{\mathrm{rank}}
&=
-\frac{1}{|\mathcal{P}|}
\sum_{(i,j)\in\mathcal{P}}
\log\sigma(R_i-R_j).
\end{aligned}
\label{eq:human_objective}
\end{equation}
We set $\mathcal{L}_{\mathrm{rank}}=0$ when no valid pairs are available. Regression fits the human rating scale, while the ranking term penalizes incorrect relative orderings. Only the cross-modal Transformer, its learnable tokens, and the score head are updated. Training settings are reported in Section~\ref{sec:experimental-setup}.
\section{Experiments}

\subsection{Experimental Setup}
\label{sec:experimental-setup}
\definecolor{mypaleblue}{rgb}{0.85, 0.92, 1.0}
\begin{table}[t]
\centering
\caption{\textbf{SyncReward as
a metric.} We report Spearman's rank correlation ($\rho\times100$) and pairwise
ranking accuracy (Pair Acc., \%) for four on-screen sound categories: Speech, Onset, Instrument, and
Ambient.
Clips are grouped by the sound category used to construct their prompts.
All comparisons are rank-based and require no score calibration.
The best and second-best results are shown in \textbf{bold} and
\underline{underlined}, respectively.
The Instrument subset is smaller because music-dominant clips were
removed from the evaluation set.}
\label{tab:metric-category}
\setlength{\tabcolsep}{3pt}
\footnotesize
\resizebox{\textwidth}{!}{%
\begin{tabular}{lccccccccc}
\toprule
\multirow{2}{*}{Method} & \multirow{2}{*}{Params} & \multicolumn{2}{c}{\textbf{Speech}} & \multicolumn{2}{c}{\textbf{Onset}} & \multicolumn{2}{c}{\textbf{Instrument}} & \multicolumn{2}{c}{\textbf{Ambient}} \\
\cmidrule(lr){3-4} \cmidrule(lr){5-6} \cmidrule(lr){7-8} \cmidrule(lr){9-10}
& & Spearman $\rho$ & Pair Acc & Spearman $\rho$ & Pair Acc & Spearman $\rho$ & Pair Acc & Spearman $\rho$ & Pair Acc \\
\midrule
\rowcolor{gray!20}\multicolumn{10}{l}{\emph{Synchronization metrics}} \\
AV-Align~\citep{yariv2024avalign}                & -- & 4.4 & 51.8 & 7.0 & 47.3 & 14.9 & 43.2 & 15.3 & 55.9 \\
CAVP~\citep{luo2023cavp}                         & 114M & 31.1 & 62.2 & 36.2 & 64.2 & 42.5 & 66.9 & 43.7 & 67.4 \\
SparseSelector~\citep{iashin2022sparseselector}  & 55M & 47.2 & 65.5 & 48.3 & 68.1 & 48.4 & 68.9 & 53.1 & 69.3 \\
Synchformer~\citep{iashin2024synchformer}        & 237M & 32.9 & 60.4 & 47.3 & 67.9 & 32.0 & 61.9 & 56.5 & 70.7 \\
PEAVS~\citep{goncalves2024peavs}                 & 108M & 16.5 & 56.5 & 1.6 & 50.6 & 5.0 & 47.9 & 12.7 & 45.0 \\
\rowcolor{gray!20}\multicolumn{10}{l}{\emph{Omni-LLM based}} \\
Qwen3-Omni-30B-A3B                                & 30BA3B & 36.3 & 64.6 & 26.8 & 60.6 & 23.6 & 59.1 & 52.4 & 71.2 \\
Qwen2.5-Omni-3B                                   & 3B & 23.9 & 59.5 & 26.0 & 60.2 & 29.8 & 61.5 & 43.7 & 67.3 \\
SynthSync~\citep{qian2026timeshiftsadaptingomnillm} & 3B & 4.9 & 51.9 & 3.1 & 51.2 & 21.9 & 58.5 & 15.1 & 55.7 \\
Qwen2.5-Omni-3B-FT                                & 3B & 61.9 & 75.7 & 64.7 & 76.6 & 51.8 & 71.0 & 73.9 & 81.3 \\
Qwen3-Omni-30B-A3B-FT                             & 30BA3B & 69.5 & 79.4 & 68.6 & 78.8 & 64.7 & 77.1 & 77.2 & 82.9 \\
\rowcolor{gray!20}\multicolumn{10}{l}{\emph{Ours}} \\
\rowcolor{mypaleblue}\textbf{SyncReward}          & 1.6B & \textbf{79.7} & \textbf{85.1} & \textbf{70.5} & \textbf{79.9} & \textbf{68.1} & \textbf{79.3} & \textbf{82.1} & \textbf{85.9} \\
\bottomrule
\end{tabular}%
}
\end{table}
\vspace{-6pt}

\paragraph{Base Model and Training Data.}
We initialize SyncReward from the Perception Encoder Audio--Visual checkpoint (\texttt{pe-av-large})~\citep{vyas2025peav}. For Stage~1, we perform semantic--temporal contrastive learning on AudioSet~\citep{audioset} and VGGSound~\citep{chen2020vggsound}, which contain totally $2.2$M natural audio--visual clips. For Stage~2, we train the human-aligned reward model on \texttt{SyncReward-Data-80K}, which comprises $80$K generated audio--visual clips with professional ordinal synchronization ratings. The clips are produced by the eight proprietary and open-source audio--visual generators described in Section~\ref{sec:data}.

\paragraph{Baselines.}
We compare SyncReward against two groups of baselines. First, we consider synchronization-specific metrics, including AV-Align~\citep{yariv2024avalign}, CAVP~\citep{luo2023cavp}, SparseSelector~\citep{iashin2022sparseselector}, Synchformer~\citep{iashin2024synchformer}, and PEAVS~\citep{goncalves2024peavs}. Second, we evaluate general-purpose omni-modal LLM judges, including Qwen2.5-Omni-3B~\citep{Qwen2.5-Omni}, Qwen3-Omni-30B-A3B~\citep{xu2025qwen3omnitechnicalreport}, and SynthSync~\citep{qian2026timeshiftsadaptingomnillm}. To construct stronger data-matched baselines, we additionally fine-tune Qwen2.5-Omni-3B and Qwen3-Omni-30B-A3B on the same \texttt{SyncReward-Data-80K}. 

\paragraph{Implementation Details.}
In Stage~1, each video is represented by $18$ overlapping segments, each spanning $0.5$\,s with a stride of $0.25$\,s. We optimize the audio and visual encoders using AdamW with a learning rate of $1\times10^{-5}$ on $4\times8$ NVIDIA H20 GPUs and a per-GPU batch size of $2$. In Stage~2, we freeze both modality encoders and train only the cross-modal interaction Transformer and scalar reward head. The reward head produces $R(x)=2\sigma(\cdot)\in[0,2]$. We optimize on $2\times8$ NVIDIA H20 GPUs, using a per-GPU batch size of $8$.

\paragraph{Evaluation Protocol.}
We evaluate SyncReward on the proposed \texttt{SyncReward-Bench},
both as a synchronization metric and as a reward for improving audio--visual generation. For metric evaluation, we report Spearman's rank correlation $\rho$ and pairwise ranking accuracy. Spearman's $\rho$ measures monotonic agreement with human ratings. For pairwise accuracy, we construct clip pairs with different mean human ratings and treat the higher-rated clip as the preferred one; a prediction is correct if the metric recovers this ordering. For reward evaluation, we apply SyncReward to Best-of-16 selection and DiffusionNFT~\citep{zheng2026diffusionnft} post-training. The resulting videos are evaluated using human and Gemini pairwise preferences together with four automatic synchronization metrics. Detailed protocols for the human evaluation and Gemini judge are provided in the appendix.

\subsection{SyncReward as a Synchronization Metric}
\label{sec:metric}

Table~\ref{tab:metric-category} compares all methods with human synchronization ratings across four sound categories in \texttt{SyncReward-Bench}. SyncReward achieves the highest Spearman correlation and pairwise accuracy in every category. Compared with the strongest baseline, Qwen3-Omni-30B-A3B-FT, it improves Spearman correlation by $1.9$--$10.2$ points and pairwise accuracy by $1.1$--$5.7$ points, with the largest margin on Speech. Methods trained mainly on real videos transfer less reliably to generated content: AV-Align, for example, falls below $50\%$ pairwise accuracy on Onset and Instrument, while the other specialized metrics show only moderate agreement with human ratings. Fine-tuning the Qwen-Omni models on the same human-rated data produces large gains, showing that generated-video failures differ from general-purpose pretraining. Even with the same training data, both Qwen baselines remain below the smaller SyncReward model across all four categories.

\subsection{SyncReward as a Reward}
\label{sec:reward}

\definecolor{mypaleblue}{rgb}{0.85, 0.92, 1.0}

\begin{table}[t]
\centering
\caption{\textbf{SyncReward as a reward.}
We evaluate synchronization rewards for inference-time Best-of-16 selection and DiffusionNFT post-training. Human and Gemini preference scores are measured on matched held-out prompts, while complementary automatic metrics are computed on the same generated videos. The current preference scores are provisional estimates from base-referenced comparisons and will be replaced by complete all-pairs evaluation.}
\label{tab:reward}

\setlength{\tabcolsep}{5pt}
\renewcommand{\arraystretch}{0.94}
\footnotesize

\resizebox{0.92\textwidth}{!}{%
\begin{tabular}{lcccccc}
\toprule
\multirow{2}{*}{Method}
& \multicolumn{2}{c}{\textbf{Pairwise Preference Score (\%)}}
& \multicolumn{4}{c}{\textbf{Automatic Metrics}} \\
\cmidrule(lr){2-3}
\cmidrule(lr){4-7}
& Human $\uparrow$
& Gemini $\uparrow$
& SyncR. $\uparrow$
& DeSync $\downarrow$
& AV-Align $\uparrow$
& JavisScore $\uparrow$ \\
\midrule

LTX~2.3 (Base Model)
& 32.31
& 47.97
& 0.935
& 0.327
& 0.306
& 0.191 \\
\midrule
\rowcolor{gray!20}
\multicolumn{7}{l}{\emph{Inference-time Best-of-16 selection}} \\
AV-Align
& 36.15
& 40.83
& 0.953
& 0.300
& \textbf{0.448}
& 0.184 \\

Synchformer
& 38.90
& 41.16
& 0.985
& \textbf{0.125}
& 0.293
& 0.192 \\

SynthSync
& \underline{52.57}
& \underline{54.20}
& \underline{0.991}
& 0.267
& 0.292
& \underline{0.197} \\

\rowcolor{mypaleblue}
\textbf{SyncReward}
& \textbf{63.67}
& \textbf{71.94}
& \textbf{1.269}
& \underline{0.205}
& \underline{0.313}
& \textbf{0.207} \\

\midrule

\rowcolor{gray!20}
\multicolumn{7}{l}{\emph{DiffusionNFT post-training}} \\


AV-Align
& 40.73
& 37.89
& 0.897
& 0.380
& \textbf{0.420}
& 0.185 \\

Synchformer
& 45.03
& 35.17
& \underline{1.053}
& \textbf{0.185}
& 0.299
& \underline{0.220} \\

SynthSync
& \underline{53.71}
& \underline{52.63}
& 0.885
& 0.417
& 0.311
& 0.189 \\

\rowcolor{mypaleblue}
\textbf{SyncReward}
& \textbf{65.78}
& \textbf{67.86}
& \textbf{1.129}
& \underline{0.187}
& \underline{0.323}
& \textbf{0.223} \\

\bottomrule
\end{tabular}%
}
\end{table}
\paragraph{Setting.}
We evaluate each reward in two settings. For inference-time selection, we sample $16$ LTX-2.3 candidates for each prompt and retain the candidate with the highest reward. For post-training, we use each reward as the optimization signal to fine-tune LTX-2.3 with DiffusionNFT~\citep{zheng2026diffusionnft}. Further implementation and evaluation details are provided in the Appendix.

\paragraph{Inference-Time Selection.}
Table~\ref{tab:reward} reports the results of Best-of-16 selection. SyncReward achieves human and Gemini preference scores of $63.67$ and $71.94$, exceeding the next-best reward by $11.1$ and $17.7$ points, respectively. It also ranks first on SyncR. and JavisScore and second on DeSync and AV-Align. By comparison, AV-Align and Synchformer perform best on the metrics they directly target but receive much lower preference scores. Improvements on a specialized synchronization metric therefore do not necessarily identify the candidates preferred by human evaluators.

\paragraph{Post-Training.}
As shown in Table~\ref{tab:reward}, SyncReward remains the strongest reward after DiffusionNFT~\citep{zheng2026diffusionnft} post-training, leading the next-best method by $12.1$ points in human preference and $15.2$ points under the Gemini judge. It also achieves the best SyncReward Score and JavisScore and ranks second on DeSync and AV-Align. AV-Align post-training raises its own metric from $0.306$ to $0.420$, but degrades the other three automatic metrics relative to the base model. Synchformer similarly obtains the lowest DeSync while producing substantially lower preference scores.
The consistent results across selection and post-training show that SyncReward provides a more reliable optimization signal than other models.

\subsection{Ablation and Analysis}
\label{sec:ablation}
\definecolor{ablationgray}{RGB}{232,232,232}
\definecolor{ablationblue}{RGB}{225,237,250}

\begin{wraptable}{r}{0.54\textwidth}
\vspace{-0.55\baselineskip}
\centering
\captionsetup{font=footnotesize,skip=3pt}
\caption{\textbf{Key ablations of SyncReward.}
We evaluate the contributions of the training stages and the
semantic-temporal contrastive objective on SyncReward-Bench. Additional design ablations are reported in the Appendix.}
\label{tab:ablation}
\footnotesize
\setlength{\tabcolsep}{3pt}
\renewcommand{\arraystretch}{1.10}
\begin{tabularx}{\linewidth}{>{\raggedright\arraybackslash}Xcc}
\toprule
\textbf{Variant}
& \textbf{Spearman $\rho \uparrow$}
& \textbf{Pair Acc. (\%) $\uparrow$} \\
\midrule
\rowcolor{ablationgray}
\multicolumn{3}{l}{\emph{Training stages}} \\
PE-AV initialization
& 26.6 & 60.4 \\
Stage~1 only
& 60.5 & 74.6 \\
Stage~2 only
& 40.9 & 66.3 \\
\midrule
\rowcolor{ablationgray}
\multicolumn{3}{l}{\emph{Stage~1 ablation}} \\
w/o semantic & 74.1 & 81.4\\
w/o temporal & 70.6 & 78.9 \\

\midrule
\rowcolor{ablationgray}
\multicolumn{3}{l}{\emph{Stage~2 ablation}} \\

w/o regression loss& 75.4& 82.0 \\

w/o rank loss& 73.8& 79.8 \\
\midrule
\rowcolor{ablationblue}
\textbf{SyncReward}
& \textbf{78.2}
& \textbf{83.6} \\
\bottomrule
\end{tabularx}
\vspace{-0.35\baselineskip}
\end{wraptable}
\paragraph{Ablation Study.} Table~\ref{tab:ablation} examines the contributions of the training stages and objectives. The full model outperforms both single-stage variants, with human alignment improving Spearman correlation by $17.7$ points and pairwise accuracy by $9.0$ points over Stage~1 alone. Training Stage~2 directly from PE-AV yields substantially lower performance, supporting the importance of temporal pretraining before human alignment. Within Stage~1, removing temporal negatives causes a larger correlation drop than removing semantic negatives ($7.6$ versus $4.1$ points). This comparison supports learning temporal correspondence beyond semantic similarity, while retaining both types of negatives gives the best results. In Stage~2, removing either regression or ranking supervision reduces performance. Omitting the ranking loss lowers pairwise accuracy by $3.8$ points, compared with $1.6$ points without regression, supporting explicit ranking supervision even when scalar ratings are available.
\begin{figure*}[t]
    \centering
    \includegraphics[width=\textwidth]{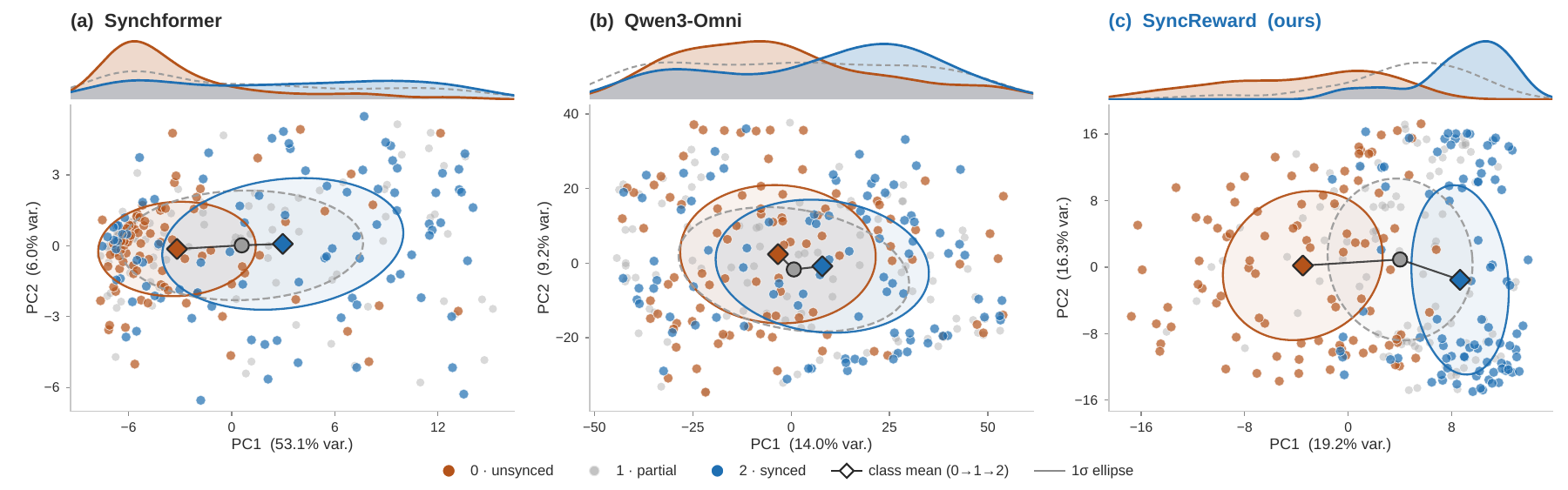}
    \caption{\textbf{Clip representations and human synchronization ratings.} Each panel shows a separate PCA projection fitted on validation features, with a class-balanced subset of held-out test clips colored by human synchronization ratings. Curves show class-wise PC1 densities; ellipses indicate one-standard-deviation covariance contours, and connected markers denote class means in rating order. Axis labels report explained variance, and coordinates are not shared across models.}
    \label{fig:analysis}
\end{figure*}

\paragraph{Representation Analysis.}
We examine whether the dominant variation in clip representations reflects human synchronization judgments. We fit PCA on validation features and apply the projections to held-out test clips (Figure~\ref{fig:analysis}). Correlations are computed on all 1,000 test clips using unmodified representations. SyncReward's first principal component has a higher absolute Spearman correlation with human ratings ($0.47$) than those of Synchformer ($0.29$) and Qwen3-Omni ($0.04$). A linear probe trained on validation features also achieves higher test correlation with SyncReward features ($0.41$ versus $0.28$ and $0.25$). These results suggest that synchronization information is more linearly accessible in SyncReward's representation. However, Qwen3-Omni's low PC1 correlation does not rule out such information in other dimensions.

\subsection{Case Study}
\label{sec:case-study}

\begin{figure}[t]
    \centering
    \includegraphics[width=\textwidth]{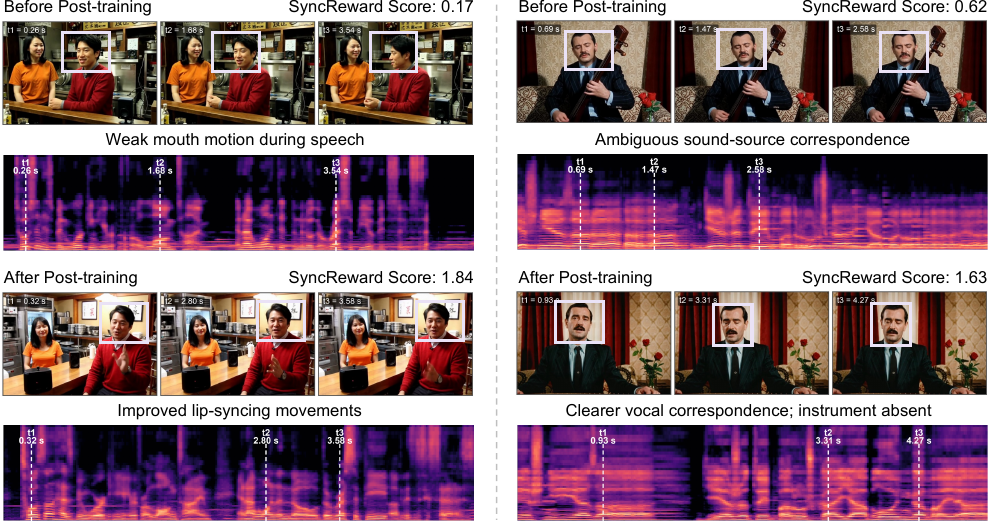}
    \caption{\textbf{Qualitative effects of SyncReward-guided DiffusionNFT post-training.} Each example compares outputs before (top) and after (bottom) post-training using three video frames and the corresponding log-mel spectrogram. Dashed lines mark frame timestamps $t_1$--$t_3$, and displayed scores are SyncReward predictions. Left: Post-training improves visible lip articulation during speech. Right: Vocal correspondence improves, but the instrument specified by the prompt is absent, illustrating a trade-off between synchronization and prompt fidelity.}
    \label{fig:case-study}
\end{figure}

Figure~\ref{fig:case-study} presents two qualitative examples of SyncReward-guided DiffusionNFT post-training. In the dialogue example, post-training produces clearer mouth articulation during speech, while the SyncReward score increases from $0.17$ to $1.84$. The singing example shows clearer correspondence between vocal audio and the visible speaker, with the score increasing from $0.62$ to $1.63$. However, the post-trained output omits the instrument specified by the prompt, highlighting a potential trade-off between synchronization and prompt fidelity. This example motivates combining synchronization rewards with complementary objectives that preserve the requested scene content. Additional qualitative comparisons are available on our anonymous demo page.

\section{Conclusion}
In this paper, we introduce \textbf{SyncReward-Data-80K}, \textbf{SyncReward-Bench}, and \textbf{SyncReward} to support the evaluation and improvement of audio--visual synchronization. The dataset and disjoint benchmark provide professional synchronization annotations across diverse sound events and generators. Building on these resources, SyncReward combines temporal pretraining on real clips with human supervision on generated clips. As an evaluation metric, it achieves the strongest agreement with human ratings across all four sound-event families among the evaluated methods. As a reward, it yields higher human preference scores than the evaluated alternatives in both best-of-$N$ selection and reinforcement learning based post-training. These results highlight the value of human-annotated generated content for learning effective synchronization rewards. Our case studies also show that optimizing synchronization alone can reduce prompt fidelity, motivating complementary objectives that preserve the intended content. We will release the data, benchmark, model, and code to support further research on audio--visual generation.

\section*{AI Use Statement}

Generative AI tools were used solely for language polishing and improving
the clarity of the manuscript. They were not used to generate experimental
results, conduct data analysis, or make scientific conclusions. All content
was reviewed and verified by the authors.

\section*{Reproducibility Statement}

We provide details of data construction, human annotation, model architecture,
training, and evaluation in the main paper and appendix. Anonymous source code
is provided with the submission, and the data, benchmark, model, and code will
be released upon publication.

\section*{Ethics Statement}

Professional annotators evaluated the synchronization quality of generated
audio--visual clips and were compensated for their participation. No sensitive
personal information was collected. The evaluated content was screened to
avoid harmful or inappropriate material.
\bibliography{main}

@inproceedings{DBLP:conf/eccv/QianWXWFLSLW26,
  author       = {Yijie Qian and
                  Juncheng Wang and
                  Chao Xu and
                  Huihan Wang and
                  Yuxiang Feng and
                  Yang Liu and
                  Baigui Sun and
                  Yong Liu and
                  Shujun Wang},
  editor       = {Paolo Favaro and
                  Zuzana Kukelova and
                  Atsuto Maki and
                  Anna Rohrbach and
                  Konrad Schindler and
                  Federico Tombari},
  title        = {Beyond Time Shifts: Adapting Omni-LLM as a Reference-Free Evaluator
                  for Generative Audio-Visual Models},
  booktitle    = {Computer Vision - {ECCV} 2026 - 19th European Conference, Malm{\"{o}},
                  Sweden, September 8-12, 2026, Proceedings, Part {LV}},
  series       = {Lecture Notes in Computer Science},
  volume       = {17055},
  pages        = {39--58},
  publisher    = {Springer},
  year         = {2026},
  url          = {https://doi.org/10.1007/978-3-032-37261-1\_3},
  doi          = {10.1007/978-3-032-37261-1\_3},
  bibsource    = {dblp computer science bibliography, https://dblp.org}
}

@article{DBLP:journals/corr/abs-2604-08540,
  author       = {Ziwei Zhou and
                  Zeyuan Lai and
                  Rui Wang and
                  Yifan Yang and
                  Zhen Xing and
                  Yuqing Yang and
                  Qi Dai and
                  Lili Qiu and
                  Chong Luo},
  title        = {AVGen-Bench: {A} Task-Driven Benchmark for Multi-Granular Evaluation
                  of Text-to-Audio-Video Generation},
  journal      = {CoRR},
  volume       = {abs/2604.08540},
  year         = {2026},
  url          = {https://doi.org/10.48550/arXiv.2604.08540},
  doi          = {10.48550/ARXIV.2604.08540},
  eprinttype   = {arXiv},
  eprint       = {2604.08540},
  bibsource    = {dblp computer science bibliography, https://dblp.org}
}

@article{DBLP:journals/corr/abs-2512-09299,
  author       = {Daili Hua and
                  Xizhi Wang and
                  Bohan Zeng and
                  Xinyi Huang and
                  Hao Liang and
                  Junbo Niu and
                  Xinlong Chen and
                  Quanqing Xu and
                  Wentao Zhang},
  title        = {VABench: {A} Comprehensive Benchmark for Audio-Video Generation},
  journal      = {CoRR},
  volume       = {abs/2512.09299},
  year         = {2025},
  url          = {https://doi.org/10.48550/arXiv.2512.09299},
  doi          = {10.48550/ARXIV.2512.09299},
  eprinttype   = {arXiv},
  eprint       = {2512.09299},
  bibsource    = {dblp computer science bibliography, https://dblp.org}
}

@article{DBLP:journals/corr/abs-2503-23377,
  author       = {Kai Liu and
                  Wei Li and
                  Lai Chen and
                  Shengqiong Wu and
                  Yanhao Zheng and
                  Jiayi Ji and
                  Fan Zhou and
                  Rongxin Jiang and
                  Jiebo Luo and
                  Hao Fei and
                  Tat{-}Seng Chua},
  title        = {JavisDiT: Joint Audio-Video Diffusion Transformer with Hierarchical
                  Spatio-Temporal Prior Synchronization},
  journal      = {CoRR},
  volume       = {abs/2503.23377},
  year         = {2025},
  url          = {https://doi.org/10.48550/arXiv.2503.23377},
  doi          = {10.48550/ARXIV.2503.23377},
  eprinttype   = {arXiv},
  eprint       = {2503.23377},
  bibsource    = {dblp computer science bibliography, https://dblp.org}
}

@article{DBLP:journals/corr/abs-2412-15159,
  author       = {Jiacheng Zhang and
                  Jie Wu and
                  Weifeng Chen and
                  Yatai Ji and
                  Xuefeng Xiao and
                  Weilin Huang and
                  Kai Han},
  title        = {OnlineVPO: Align Video Diffusion Model with Online Video-Centric Preference
                  Optimization},
  journal      = {CoRR},
  volume       = {abs/2412.15159},
  year         = {2024},
  url          = {https://doi.org/10.48550/arXiv.2412.15159},
  doi          = {10.48550/ARXIV.2412.15159},
  eprinttype   = {arXiv},
  eprint       = {2412.15159},
  bibsource    = {dblp computer science bibliography, https://dblp.org}
}

@inproceedings{DBLP:conf/cvpr/LiuWZWHPC25,
  author       = {Runtao Liu and
                  Haoyu Wu and
                  Ziqiang Zheng and
                  Chen Wei and
                  Yingqing He and
                  Renjie Pi and
                  Qifeng Chen},
  title        = {VideoDPO: Omni-Preference Alignment for Video Diffusion Generation},
  booktitle    = {{IEEE/CVF} Conference on Computer Vision and Pattern Recognition,
                  {CVPR} 2025, Nashville, TN, USA, June 11-15, 2025},
  pages        = {8009--8019},
  publisher    = {Computer Vision Foundation / {IEEE}},
  year         = {2025},
  url          = {https://openaccess.thecvf.com/content/CVPR2025/html/Liu\_VideoDPO\_Omni-Preference\_Alignment\_for\_Video\_Diffusion\_Generation\_CVPR\_2025\_paper.html},
  doi          = {10.1109/CVPR52734.2025.00750},
  bibsource    = {dblp computer science bibliography, https://dblp.org}
}

@article{DBLP:journals/corr/abs-2412-04814,
  author       = {Yibin Wang and
                  Zhiyu Tan and
                  Junyan Wang and
                  Xiaomeng Yang and
                  Cheng Jin and
                  Hao Li},
  title        = {LiFT: Leveraging Human Feedback for Text-to-Video Model Alignment},
  journal      = {CoRR},
  volume       = {abs/2412.04814},
  year         = {2024},
  url          = {https://doi.org/10.48550/arXiv.2412.04814},
  doi          = {10.48550/ARXIV.2412.04814},
  eprinttype   = {arXiv},
  eprint       = {2412.04814},
  bibsource    = {dblp computer science bibliography, https://dblp.org}
}

@inproceedings{DBLP:conf/nips/LiFFWBCW24,
  author       = {Jiachen Li and
                  Weixi Feng and
                  Tsu{-}Jui Fu and
                  Xinyi Wang and
                  Sugato Basu and
                  Wenhu Chen and
                  William Yang Wang},
  editor       = {Amir Globersons and
                  Lester Mackey and
                  Danielle Belgrave and
                  Angela Fan and
                  Ulrich Paquet and
                  Jakub M. Tomczak and
                  Cheng Zhang},
  title        = {T2V-Turbo: Breaking the Quality Bottleneck of Video Consistency Model
                  with Mixed Reward Feedback},
  booktitle    = {Advances in Neural Information Processing Systems 37: Annual Conference
                  on Neural Information Processing Systems 2024, NeurIPS 2024, Vancouver,
                  BC, Canada, December 10 - 15, 2024},
  year         = {2024},
  url          = {http://papers.nips.cc/paper\_files/paper/2024/hash/8a57aa8e8b57e64a42e95f7dceb0adb9-Abstract-Conference.html},
  bibsource    = {dblp computer science bibliography, https://dblp.org}
}

@article{DBLP:journals/corr/abs-2407-08737,
  author       = {Mihir Prabhudesai and
                  Russell Mendonca and
                  Zheyang Qin and
                  Katerina Fragkiadaki and
                  Deepak Pathak},
  title        = {Video Diffusion Alignment via Reward Gradients},
  journal      = {CoRR},
  volume       = {abs/2407.08737},
  year         = {2024},
  url          = {https://doi.org/10.48550/arXiv.2407.08737},
  doi          = {10.48550/ARXIV.2407.08737},
  eprinttype   = {arXiv},
  eprint       = {2407.08737},
  bibsource    = {dblp computer science bibliography, https://dblp.org}
}

@inproceedings{DBLP:conf/cvpr/YuanZW0FPZ0A024,
  author       = {Hangjie Yuan and
                  Shiwei Zhang and
                  Xiang Wang and
                  Yujie Wei and
                  Tao Feng and
                  Yining Pan and
                  Yingya Zhang and
                  Ziwei Liu and
                  Samuel Albanie and
                  Dong Ni},
  title        = {InstructVideo: Instructing Video Diffusion Models with Human Feedback},
  booktitle    = {{IEEE/CVF} Conference on Computer Vision and Pattern Recognition,
                  {CVPR} 2024, Seattle, WA, USA, June 16-22, 2024},
  pages        = {6463--6474},
  publisher    = {{IEEE}},
  year         = {2024},
  url          = {https://doi.org/10.1109/CVPR52733.2024.00618},
  doi          = {10.1109/CVPR52733.2024.00618},
  bibsource    = {dblp computer science bibliography, https://dblp.org}
}

@inproceedings{DBLP:conf/nips/LiuLLLLWWZO25,
  author       = {Jie Liu and
                  Gongye Liu and
                  Jiajun Liang and
                  Yangguang Li and
                  Jiaheng Liu and
                  Xintao Wang and
                  Pengfei Wan and
                  Di Zhang and
                  Wanli Ouyang},
  editor       = {Danielle Belgrave and
                  Cheng Zhang and
                  Laura N. Montoya and
                  Hsuan{-}Tien Lin and
                  Razvan Pascanu and
                  Piotr Koniusz and
                  Marzyeh Ghassemi and
                  Nancy Chen and
                  Iv{\'{a}}n Vladimir Meza Ru{\'{\i}}z and
                  Arturo Loaiza{-}Bonilla},
  title        = {Flow-GRPO: Training Flow Matching Models via Online {RL}},
  booktitle    = {Advances in Neural Information Processing Systems 38: Annual Conference
                  on Neural Information Processing Systems 2025, NeurIPS 2025, San Diego,
                  CA, USA, December 2-7, 2025 / Mexico City, Mexico, November 30 - December
                  5, 2025},
  year         = {2025},
  url          = {http://papers.nips.cc/paper\_files/paper/2025/hash/3a10c46572628d58cb44fb705f25cbbf-Abstract-Conference.html},
  bibsource    = {dblp computer science bibliography, https://dblp.org}
}

@article{DBLP:journals/corr/abs-2505-07818,
  author       = {Zeyue Xue and
                  Jie Wu and
                  Yu Gao and
                  Fangyuan Kong and
                  Lingting Zhu and
                  Mengzhao Chen and
                  Zhiheng Liu and
                  Wei Liu and
                  Qiushan Guo and
                  Weilin Huang and
                  Ping Luo},
  title        = {DanceGRPO: Unleashing {GRPO} on Visual Generation},
  journal      = {CoRR},
  volume       = {abs/2505.07818},
  year         = {2025},
  url          = {https://doi.org/10.48550/arXiv.2505.07818},
  doi          = {10.48550/ARXIV.2505.07818},
  eprinttype   = {arXiv},
  eprint       = {2505.07818},
  bibsource    = {dblp computer science bibliography, https://dblp.org}
}

@inproceedings{DBLP:conf/nips/LiuLLYLZWWXWLYW25,
  author       = {Jie Liu and
                  Gongye Liu and
                  Jiajun Liang and
                  Ziyang Yuan and
                  Xiaokun Liu and
                  Mingwu Zheng and
                  Xiele Wu and
                  Qiulin Wang and
                  Menghan Xia and
                  Xintao Wang and
                  Xiaohong Liu and
                  Fei Yang and
                  Pengfei Wan and
                  Di Zhang and
                  Kun Gai and
                  Yujiu Yang and
                  Wanli Ouyang},
  editor       = {Danielle Belgrave and
                  Cheng Zhang and
                  Laura N. Montoya and
                  Hsuan{-}Tien Lin and
                  Razvan Pascanu and
                  Piotr Koniusz and
                  Marzyeh Ghassemi and
                  Nancy Chen and
                  Iv{\'{a}}n Vladimir Meza Ru{\'{\i}}z and
                  Arturo Loaiza{-}Bonilla},
  title        = {Improving Video Generation with Human Feedback},
  booktitle    = {Advances in Neural Information Processing Systems 38: Annual Conference
                  on Neural Information Processing Systems 2025, NeurIPS 2025, San Diego,
                  CA, USA, December 2-7, 2025 / Mexico City, Mexico, November 30 - December
                  5, 2025},
  year         = {2025},
  url          = {http://papers.nips.cc/paper\_files/paper/2025/hash/76227feb18ea0ee40bd15cf02c33e18e-Abstract-Conference.html},
  bibsource    = {dblp computer science bibliography, https://dblp.org}
}

@misc{kling,
  title={Kling Video Generation},
  author={Kuaishou, Inc},
  year={2024},
  url={https://klingai.com/},
}

@inproceedings{yariv2024avalign,
  title={Diverse and aligned audio-to-video generation via text-to-video model adaptation},
  author={Yariv, Guy and Gat, Itai and Benaim, Sagie and Wolf, Lior and Schwartz, Idan and Adi, Yossi},
  booktitle={Proceedings of the AAAI Conference on Artificial Intelligence},
  volume={38},
  pages={6639--6647},
  year={2024}
}

@article{Qwen2.5-Omni,
  title={Qwen2.5-Omni Technical Report},
    author={Jin Xu and Zhifang Guo and Jinzheng He and Hangrui Hu and Ting He and Shuai Bai and Keqin Chen and Jialin Wang and Yang Fan and Kai Dang and Bin Zhang and Xiong Wang and Yunfei Chu and Junyang Lin},
  journal={arXiv preprint arXiv:2503.20215},
  year={2025}
}

@article{hacohen2026ltx,
  title={LTX-2: Efficient Joint Audio-Visual Foundation Model},
  author={HaCohen, Yoav and Brazowski, Benny and Chiprut, Nisan and Bitterman, Yaki and Kvochko, Andrew and Berkowitz, Avishai and Shalem, Daniel and Lifschitz, Daphna and Moshe, Dudu and Porat, Eitan and others},
  journal={arXiv preprint arXiv:2601.03233},
  year={2026}
}

@misc{low2025ovi,
      title={Ovi: Twin Backbone Cross-Modal Fusion for Audio-Video Generation}, 
      author={Chetwin Low and Weimin Wang and Calder Katyal},
      year={2025},
      eprint={2510.01284},
      archivePrefix={arXiv},
      primaryClass={cs.MM},
      url={https://arxiv.org/abs/2510.01284}, 
}

@misc{guan2026bridgedit,
      title={Taming Text-to-Sounding Video Generation via Advanced Modality Condition and Interaction}, 
      author={Kaisi Guan and Xihua Wang and Zhengfeng Lai and Xin Cheng and Peng Zhang and XiaoJiang Liu and Ruihua Song and Meng Cao},
      year={2026},
      eprint={2510.03117},
      archivePrefix={arXiv},
      primaryClass={cs.CV},
      url={https://arxiv.org/abs/2510.03117}, 
}

@misc{openmoss_mova_2026,
  title         = {MOVA: Towards Scalable and Synchronized Video-Audio Generation},
  author        = {{SII-OpenMOSS Team} and Donghua Yu and Mingshu Chen and Qi Chen and Qi Luo and Qianyi Wu and Qinyuan Cheng and Ruixiao Li and Tianyi Liang and Wenbo Zhang and Wenming Tu and Xiangyu Peng and Yang Gao and Yanru Huo and Ying Zhu and Yinze Luo and Yiyang Zhang and Yuerong Song and Zhe Xu and Zhiyu Zhang and Chenchen Yang and Cheng Chang and Chushu Zhou and Hanfu Chen and Hongnan Ma and Jiaxi Li and Jingqi Tong and Junxi Liu and Ke Chen and Shimin Li and Songlin Wang and Wei Jiang and Zhaoye Fei and Zhiyuan Ning and Chunguo Li and Chenhui Li and Ziwei He and Zengfeng Huang and Xie Chen and Xipeng Qiu},
  year          = {2026},
  month         = feb,
  eprint        = {2602.08794},
  archivePrefix = {arXiv},
  primaryClass  = {cs.CV},
  doi           = {10.48550/arXiv.2602.08794},
  url           = {https://arxiv.org/abs/2602.08794},
  note          = {Technical report. Corresponding authors: Xie Chen and Xipeng Qiu. Project leaders: Qinyuan Cheng and Tianyi Liang.}
}

@misc{ji2026nava,
      title         = {Native Audio-Visual Alignment for Generation},
      author        = {Longbin Ji and Guan Wang and Xuan Wei and Chenye Yang and Xiangrui Liu and Zhenyu Zhang and Shuohuan Wang and Yu Sun and Jingzhou He},
      year          = {2026},
      eprint        = {2605.30073},
      archivePrefix = {arXiv},
      primaryClass  = {cs.CV},
      url           = {https://arxiv.org/abs/2605.30073},
}

@article{davinci-magihuman-2026,
  title={Speed by Simplicity: A Single-Stream Architecture for Fast Audio-Video Generative Foundation Model},
  author={SII-GAIR and Sand. ai and Chern, Ethan and Teng, Hansi and Sun, Hanwen and Wang, Hao and Pan, Hong and Jia, Hongyu and Su, Jiadi and Li, Jin and Yu, Junjie and Liu, Lijie and Li, Lingzhi and Ye, Lyumanshan and Hu, Min and Wang, Qiangang and Qi, Quanwei and Chern, Steffi and Bu, Tao and Wang, Taoran and Xu, Teren and Zhang, Tianning and Mi, Tiantian and Xu, Weixian and Zhang, Wenqiang and Zhang, Wentai and Yi, Xianping and Cai, Xiaojie and Kang, Xiaoyang and Ma, Yan and Liu, Yixiu and Zhang, Yunbo and Huang, Yunpeng and Lin, Yutong and Tao, Zewei and Liu, Zhaoliang and Zhang, Zheng and Cen, Zhiyao and Yu, Zhixuan and Wang, Zhongshu and Hu, Zhulin and Zhou, Zijin and Guo, Zinan and Cao, Yue and Liu, Pengfei},
  journal={arXiv preprint arXiv:2603.21986},
  year={2026}
}

@inproceedings{
    magi2,
    title={MAGI-2 Preview: Scaling Video Generation Models Efficiently}, 
    author={Sandai},
    year={2026},
    url={https://sand.ai/blog/magi-2-preview}
}

@inproceedings{
    minimaxh3,
    title={MiniMax H3}, 
    author={MiniMaxAI},
    year={2026},
    url={https://huggingface.co/MiniMaxAI/MiniMax-H3}
}

@inproceedings{iashin2024synchformer,
  title={Synchformer: Efficient synchronization from sparse cues},
  author={Iashin, Vladimir and Xie, Weidi and Rahtu, Esa and Zisserman, Andrew},
  booktitle={ICASSP 2024-2024 IEEE International Conference on Acoustics, Speech and Signal Processing (ICASSP)},
  pages={5325--5329},
  year={2024},
  organization={IEEE}
}

@inproceedings{goncalves2024peavs,
  title={Perceptual evaluation of audio-visual synchrony grounded in viewers’ opinion scores},
  author={Goncalves, Lucas and Mathur, Prashant and Lavania, Chandrashekhar and Cekic, Metehan and Federico, Marcello and Han, Kyu J},
  booktitle={European Conference on Computer Vision},
  pages={288--305},
  year={2024},
  organization={Springer}
}

@article{iashin2022sparseselector,
  title={Sparse in space and time: Audio-visual synchronisation with trainable selectors},
  author={Iashin, Vladimir and Xie, Weidi and Rahtu, Esa and Zisserman, Andrew},
  journal={arXiv preprint arXiv:2210.07055},
  year={2022}
}

@article{luo2023cavp,
  title={Diff-foley: Synchronized video-to-audio synthesis with latent diffusion models},
  author={Luo, Simian and Yan, Chuanhao and Hu, Chenxu and Zhao, Hang},
  journal={Advances in Neural Information Processing Systems},
  volume={36},
  pages={48855--48876},
  year={2023}
}

@misc{guan2026fateframelevelaudiovisualtemporal,
      title={FATE: Frame-Level Audio-Visual Temporal Embedding}, 
      author={Kaisi Guan and Bingzi Zhang and Xihua Wang and Ying Ba and Xin Cheng and Yijing Chen and Ruihua Song},
      year={2026},
      eprint={2608.01310},
      archivePrefix={arXiv},
      primaryClass={cs.MM},
      url={https://arxiv.org/abs/2608.01310}, 
}

@misc{qian2026timeshiftsadaptingomnillm,
      title={Beyond Time Shifts: Adapting Omni-LLM as a Reference-Free Evaluator for Generative Audio-Visual Models}, 
      author={Yijie Qian and Juncheng Wang and Chao Xu and Huihan Wang and Yuxiang Feng and Yang Liu and Baigui Sun and Yong Liu and Shujun Wang},
      year={2026},
      eprint={2607.09091},
      archivePrefix={arXiv},
      primaryClass={cs.CV},
      url={https://arxiv.org/abs/2607.09091}, 
}

@inproceedings{chung2017syncnet,
  title={Out of Time: Automated Lip Sync in the Wild},
  author={Chung, Joon Son and Zisserman, Andrew},
  booktitle={Computer Vision -- ACCV 2016 Workshops},
  pages={251--263},
  year={2017},
  organization={Springer}
}

@misc{vyas2025peav,
      title={Pushing the Frontier of Audiovisual Perception with Large-Scale Multimodal Correspondence Learning}, 
      author={Apoorv Vyas and Heng-Jui Chang and Cheng-Fu Yang and Po-Yao Huang and Luya Gao and Julius Richter and Sanyuan Chen and Matt Le and Piotr Dollár and Christoph Feichtenhofer and Ann Lee and Wei-Ning Hsu},
      year={2025},
      eprint={2512.19687},
      archivePrefix={arXiv},
      primaryClass={cs.SD},
      url={https://arxiv.org/abs/2512.19687}, 
}

@inproceedings{audioset,title	= {Audio Set: An ontology and human-labeled dataset for audio events},author	= {Jort F. Gemmeke and Daniel P. W. Ellis and Dylan Freedman and Aren Jansen and Wade Lawrence and R. Channing Moore and Manoj Plakal and Marvin Ritter},year	= {2017},booktitle	= {Proc. IEEE ICASSP 2017},address	= {New Orleans, LA}}

@misc{chen2020vggsound,
      title={VGGSound: A Large-scale Audio-Visual Dataset}, 
      author={Honglie Chen and Weidi Xie and Andrea Vedaldi and Andrew Zisserman},
      year={2020},
      eprint={2004.14368},
      archivePrefix={arXiv},
      primaryClass={cs.CV},
      url={https://arxiv.org/abs/2004.14368}, 
}

@misc{seedance,
  title={Seedance Video Generation},
  author={Bytedance, Inc},
  year={2025},
  url={https://seed.bytedance.com/en/seedance2_0},
}

@misc{veo,
  title={Veo3 Video Generation},
  author={Google, Inc},
  year={2025},
  url={https://deepmind.google/models/veo/},
}

@inproceedings{zheng2026diffusionnft,
  title={Diffusionnft: Online diffusion reinforcement with forward process},
  author={Zheng, Kaiwen and Chen, Huayu and Ye, Haotian and Wang, Haoxiang and Zhang, Qinsheng and Jiang, Kai and Su, Hang and Ermon, Stefano and Zhu, Jun and Liu, Ming-Yu},
  booktitle={International Conference on Learning Representations},
  volume={2026},
  pages={134129--134150},
  year={2026}
}

@inproceedings{xu2023imagereward,
  title = {ImageReward: Learning and Evaluating Human Preferences for Text-to-Image Generation},
  author = {Xu, Jiazheng and Liu, Xiao and Wu, Yuchen and Tong, Yuxuan and Li, Qinkai and Ding, Ming and Tang, Jie and Dong, Yuxiao},
  booktitle = {Advances in Neural Information Processing Systems},
  year = {2023},
  url = {https://arxiv.org/abs/2304.05977}
}

@inproceedings{kirstain2023pickapic,
  title = {{Pick-a-Pic}: An Open Dataset of User Preferences for Text-to-Image Generation},
  author = {Kirstain, Yuval and Polyak, Adam and Singer, Uriel and Matiana, Shahbuland and Penna, Joe and Levy, Omer},
  booktitle = {Advances in Neural Information Processing Systems},
  year = {2023},
  url = {https://arxiv.org/abs/2305.01569}
}

@article{xu2024visionreward,
  title = {{VisionReward}: Fine-Grained Multi-Dimensional Human Preference Learning for Image and Video Generation},
  author = {Xu, Jiazheng and Huang, Yu and Cheng, Jiale and Yang, Yuanming and Xu, Jiajun and Wang, Yuan and Duan, Wenbo and Yang, Shen and Jin, Qunlin and Li, Shurun and Teng, Jiayan and Yang, Zhuoyi and Zheng, Wendi and Liu, Xiao and Zhang, Dan and Ding, Ming and Zhang, Xiaohan and Gu, Xiaotao and Huang, Shiyu and Huang, Minlie and Tang, Jie and Dong, Yuxiao},
  journal = {arXiv preprint arXiv:2412.21059},
  year = {2024},
  url = {https://arxiv.org/abs/2412.21059}
}

@inproceedings{wu2026editreward,
  title = {{EditReward}: A Human-Aligned Reward Model for Instruction-Guided Image Editing},
  author = {Wu, Keming and Jiang, Sicong and Ku, Max and Nie, Ping and Liu, Minghao and Chen, Wenhu},
  booktitle = {International Conference on Learning Representations},
  year = {2026},
  url = {https://openreview.net/forum?id=eZu358JOOR}
}

@inproceedings{he2024videoscore,
  title = {{VideoScore}: Building Automatic Metrics to Simulate Fine-grained Human Feedback for Video Generation},
  author = {He, Xuan and Jiang, Dongfu and Zhang, Ge and Ku, Max and Soni, Achint and Siu, Sherman and Chen, Haonan and Chandra, Abhranil and Jiang, Ziyan and Arulraj, Aaran and Wang, Kai and Do, Quy Duc and Ni, Yuansheng and Lyu, Bohan and Narsupalli, Yaswanth and Fan, Rongqi and Lyu, Zhiheng and Lin, Bill Yuchen and Chen, Wenhu},
  booktitle = {Proceedings of the 2024 Conference on Empirical Methods in Natural Language Processing},
  year = {2024},
  publisher = {Association for Computational Linguistics},
  pages = {2105--2123},
  url = {https://aclanthology.org/2024.emnlp-main.127/}
}

@article{liu2025videoreward,
  title = {Improving Video Generation with Human Feedback},
  author = {Liu, Jie and Liu, Gongye and Liang, Jiajun and Yuan, Ziyang and Liu, Xiaokun and Zheng, Mingwu and Wu, Xiele and Wang, Qiulin and Xia, Menghan and Wang, Xintao and Liu, Xiaohong and Yang, Fei and Wan, Pengfei and Zhang, Di and Gai, Kun and Yang, Yujiu and Ouyang, Wanli},
  journal = {arXiv preprint arXiv:2501.13918},
  year = {2025},
  url = {https://arxiv.org/abs/2501.13918}
}

@article{xu2025qwen3omnitechnicalreport,
      title={Qwen3-Omni Technical Report}, 
      author={Jin Xu and Zhifang Guo and Hangrui Hu and Yunfei Chu and Xiong Wang and Jinzheng He and Yuxuan Wang and Xian Shi and Ting He and Xinfa Zhu and Yuanjun Lv and Yongqi Wang and Dake Guo and He Wang and Linhan Ma and Pei Zhang and Xinyu Zhang and Hongkun Hao and Zishan Guo and Baosong Yang and Bin Zhang and Ziyang Ma and Xipin Wei and Shuai Bai and Keqin Chen and Xuejing Liu and Peng Wang and Mingkun Yang and Dayiheng Liu and Xingzhang Ren and Bo Zheng and Rui Men and Fan Zhou and Bowen Yu and Jianxin Yang and Le Yu and Jingren Zhou and Junyang Lin},
  journal = {arXiv preprint arXiv:2509.17765},
  year = {2025},
  url = {https://arxiv.org/abs/2509.17765}
}
\bibliographystyle{iclr2027_conference}

\newpage
\appendix

\section{Related Work}
\paragraph{Audio--visual Datasets and Benchmarks.}
Recent benchmarks increasingly evaluate joint audio--visual generation from multiple dimensions. JavisBench~\citep{DBLP:journals/corr/abs-2503-23377} evaluates generation quality, semantic consistency, and audio--visual spatio-temporal synchronization, while VABench~\citep{DBLP:journals/corr/abs-2512-09299} covers audio and visual quality, cross-modal consistency, synchronization, and lip--speech alignment. AVGen-Bench~\citep{DBLP:journals/corr/abs-2604-08540} further provides multi-granular evaluation spanning perceptual quality, cross-modal alignment, and fine-grained semantic controllability. These benchmarks primarily evaluate generation systems rather than provide large-scale human supervision for learning synchronization rewards. SynthSync~\citep{DBLP:conf/eccv/QianWXWFLSLW26} provides pairwise human synchronization preferences over generated audio conditioned on real videos. In contrast, SyncReward-Data-80K provides large-scale professional synchronization annotations on diverse jointly generated audio--visual outputs, while SyncReward-Bench provides an independently annotated benchmark for evaluating agreement with human synchronization judgments.

\paragraph{Post-Training for Video Generation.}
Post-training has become an important paradigm for improving video generation quality, temporal consistency, and alignment with human preferences. Early approaches directly optimize video generators with reward feedback, including InstructVideo~\citep{DBLP:conf/cvpr/YuanZW0FPZ0A024}, VADER~\citep{DBLP:journals/corr/abs-2407-08737} and T2V-Turbo~\citep{DBLP:conf/nips/LiFFWBCW24}. Another line of work learns from preference data: LiFT~\citep{DBLP:journals/corr/abs-2412-04814} adopts reward-weighted optimization, VideoDPO~\citep{DBLP:conf/cvpr/LiuWZWHPC25} extends direct preference optimization to video generation, and OnlineVPO~\citep{DBLP:journals/corr/abs-2412-15159} further performs preference optimization with samples generated by the evolving policy. With the increasing adoption of flow matching and rectified flow, recent methods have developed post-training algorithms tailored to flow-based video generators. Flow-DPO~\citep{DBLP:conf/nips/LiuLLYLZWWXWLYW25} and Flow-RWR~\citep{DBLP:conf/nips/LiuLLYLZWWXWLYW25} extend preference optimization and reward-weighted regression to rectified-flow models. DanceGRPO~\citep{DBLP:journals/corr/abs-2505-07818} and Flow-GRPO~\citep{DBLP:conf/nips/LiuLLLLWWZO25} further introduce online reinforcement learning with group-relative rewards, enabling optimization using samples generated by the current policy. More recently, DiffusionNFT~\citep{zheng2026diffusionnft} directly incorporates reward optimization into the native diffusion or flow-matching objective by contrasting high- and low-reward samples, avoiding explicit likelihood estimation over reverse-generation trajectories. 
\section{SyncReward-Data-80K Details}
\subsection{Sound-Event Taxonomy}
\label{app:taxonomy}

We organize on-screen sound events into a three-level hierarchy with four event families, 16 intermediate categories, and 58 fine-grained subcategories (Table~\ref{tab:taxonomy}). The hierarchy specifies the sound event targeted during prompt construction. Categories therefore describe the intended event rather than whether a generated clip successfully depicts it or achieves synchronization.

\paragraph{Event families.}
\textit{Speech} covers speaking and singing by visible human or animated characters, including stylized characters whose articulation may differ from human lip motion. \textit{Onset} covers action-linked sounds and discrete acoustic events, such as footsteps, impacts, animal calls, and device sounds. Here, ``Onset'' is a category name rather than a restriction to instantaneous sounds. \textit{Instrument} covers sounds produced by visible musical performances. \textit{Ambient} covers environmental sounds associated with visible sources or processes, including rain, traffic, and indoor activity. Across all four families, the target sound must be grounded in the visible scene; off-screen narration and background music are excluded.

\begin{table}[t]
\centering
\caption{\textbf{The SyncReward sound-event taxonomy.} The three levels comprise four event families, 16 intermediate categories, and 58 fine-grained subcategories. Numbers in parentheses indicate the number of subcategories in each category. Language tags are recorded separately for Speech.}
\label{tab:taxonomy}
\small
\setlength{\tabcolsep}{5pt}
\renewcommand{\arraystretch}{1.15}
\begin{tabularx}{\textwidth}{
    >{\raggedright\arraybackslash}p{0.13\textwidth}
    >{\raggedright\arraybackslash}p{0.22\textwidth}
    >{\raggedright\arraybackslash}X
}
\toprule
\textbf{Family}
& \textbf{Category}
& \textbf{Subcategories} \\
\midrule

\textbf{Speech}
& Conversation (2)
& Monologue; multi-speaker conversation. \\

& Vocal performance (3)
& Singing; expressive speech; challenging tongue twisters. \\

& Stylized characters (5)
& Japanese anime; Western cartoons; children's animation; 3D film/game characters; puppets and robots without conventional lips. \\

\midrule

\textbf{Onset}
& Human actions (4)
& Body contact; eating and chewing; breathing actions; footsteps. \\

& Object interactions (5)
& Impacts and breakage; object manipulation; material processing; liquid interactions; tool use. \\

& Mechanical and electronic events (4)
& Device startup; electronic notification sounds; vehicle operation; rhythmic mechanisms. \\

& Animals (2)
& Animal vocalizations; animal movements. \\

& Sports (5)
& Ball sports; combat sports and martial arts; athletics and gymnastics; water sports; ice and snow sports. \\

& Fictional events (4)
& Cartoon character actions; fantastical creatures; magical and energy effects; in-world game feedback. \\

\midrule

\textbf{Instrument}
& Strings (2)
& Plucked strings; bowed strings. \\

& Winds (3)
& Woodwinds; brass; free-reed instruments. \\

& Keyboards (3)
& Piano; organ; electronic keyboards. \\

& Percussion (2)
& Drums; pitched percussion. \\

\midrule

\textbf{Ambient}
& Natural processes (5)
& Rain; thunderstorms; wind; fire; flowing water. \\

& Urban and harbor environments (6)
& Street traffic; crowds; construction; aircraft; ships and boats; trains and subways. \\

& Indoor environments (3)
& Household appliances; kitchen activity; indoor crowds. \\

\bottomrule
\end{tabularx}
\end{table}

\paragraph{Language tags.}
Speech prompts additionally receive one of six language tags: English, Mandarin Chinese, Japanese, Korean, Cantonese, or Other. Language is an independent attribute rather than an additional level of the hierarchy. For example, a Japanese-anime dialogue can be tagged as English, while a monologue can be tagged as Cantonese. These tags do not increase subcategories.

\paragraph{Category interpretation.}
The taxonomy combines event type with selected scene and character attributes to support prompt coverage. Its categories are not intended to be mutually exclusive descriptions of every sound in a clip. A street scene may contain both footsteps and traffic, for example. The category used for evaluation is the target event selected during prompt construction: footsteps belong to Onset, whereas the surrounding traffic belongs to Ambient. Similarly, Speech targets vocal articulation, while Instrument targets the visible act of playing an instrument. Fictional events remain subject to the same visible-source requirement; an audible effect must correspond to an on-screen action or event.
\subsubsection{Human Annotation}
\begin{promptbox}[title={Human annotation guidelines for audio-visual synchronization}]
\ttfamily\small\raggedright
This task evaluates audio-visual temporal synchronization. For each
generated clip, annotate the audio and video in two steps.
\medskip

\textbf{\rmfamily Step 1: Is the audio related to the video?}
\begin{itemize}
\setlength{\itemsep}{1pt}
\setlength{\parskip}{0pt}
\setlength{\topsep}{2pt}
  \item 0 = Unrelated: the audio is completely unrelated to the visual content (e.g., the video shows one kind of content while the audio contains an irrelevant sound).
  \item 1 = Related: the audio semantically corresponds to the visual content, so its synchronization can be assessed.
\end{itemize}
\medskip

\textbf{\rmfamily Step 2: Synchronization score (only if Step 1 = 1):}
\begin{itemize}
\setlength{\itemsep}{1pt}
\setlength{\parskip}{0pt}
\setlength{\topsep}{2pt}
  \item 0 = Not synchronized: lip motion or actions clearly do not match the sound.
  \item 1 = Partially synchronized: audio and video are partly consistent but not aligned, with noticeable delay, lead, or unstable timing.
  \item 2 = Perfectly synchronized: the visible sound source and the audio are highly consistent in both content and timing.
\end{itemize}
\medskip

\textbf{\rmfamily Notes:}
\begin{itemize}
\setlength{\itemsep}{1pt}
\setlength{\parskip}{0pt}
\setlength{\topsep}{2pt}
  \item If Step 1 = 0, no synchronization score is given and the clip is submitted directly.
  \item A free-text comment is optional.
  \item Wear headphones and turn the volume up.
\end{itemize}
\end{promptbox}
\subsection{Prompt Construction}
\label{app:prompts}

This section describes how generation prompts are derived from source captions: taxonomy assignment, category-balanced sampling, and synchronization-oriented rewriting. All three steps use Qwen3.6-35B-A3B served with vLLM, with thinking mode disabled. The full instructions are reproduced below.

\paragraph{Source captions and taxonomy assignment.}
Source captions describe real video clips and the sounds they contain. We classify every sound mentioned in a caption into the taxonomy of Appendix~\ref{app:taxonomy} with greedy decoding (temperature $0$), processing eight captions per request. The system prompt contains the complete taxonomy guideline (Box~\ref{box:taxonomy}). For each sound, the model outputs its three taxonomy levels, whether it is on screen, a short piece of textual evidence, and a confidence level. Speech sounds additionally receive one of the six language tags. A caption may contain several independent sounds, and the model is instructed not to merge multiple sources into one class.

\paragraph{Category-balanced sampling.}
The caption pool is dominated by a few frequent categories, such as conversation and street ambience. We therefore resample captions with per-subcategory quotas that trade off taxonomy coverage against relevance for synchronization evaluation. This yields three disjoint sets of $10$K captions. Each sampled caption keeps its target subcategory, which serves as the evaluation category of every clip generated from it (Appendix~\ref{app:taxonomy}).

\paragraph{Synchronization-oriented rewriting.}
Source captions often attribute sounds to off-screen sources, for example narration, off-screen voices, or background music. Such sounds cannot be evaluated for audio--visual timing. We rewrite each sampled caption with the instruction in Box~\ref{box:rewrite} (temperature $0.7$, top-$p$ $0.8$, at most $1{,}024$ output tokens). The instruction asks the model to
(i)~enumerate every sound in the caption;
(ii)~keep on-screen sounds and state explicitly how the visible source's action corresponds to the sound, e.g., lips moving with speech or a spoon stirring with the stirring sound;
(iii)~reassign off-screen speech to a character already visible in the frame, adding a visible speaker only when no suitable character exists;
(iv)~delete sounds that cannot have a visible source, such as background music; and
(v)~preserve the original camera, setting, and character descriptions.
Rewritten prompts are written in English, while spoken lines keep their original language. This way the language tag of Speech prompts is preserved.

\paragraph{Scope.}
The prompts specify the \emph{intended} audio--visual correspondence; they do not guarantee that it is realized. Generated clips may still contain the synchronization errors that SyncReward is designed to detect. Moreover, the target subcategory records the event requested in the prompt, not the event that a generated clip actually depicts.

\begin{promptbox}[title={Taxonomy assignment prompt}]
\label{box:taxonomy}
\ttfamily\small\raggedright
\textbf{\rmfamily System:}\\
You are a sound-scene classifier for audio-visual captions. Classify the
sounds in the caption strictly following the taxonomy guideline below.
\medskip

\textbf{\rmfamily Requirements:}
\begin{itemize}
\setlength{\itemsep}{1pt}
\setlength{\parskip}{0pt}
\setlength{\topsep}{2pt}
  \item Output a single valid JSON object only; no Markdown, no explanation.
  \item If the caption contains several independent sounds, output one entry per sound; if no audio information can be inferred, output \{"sounds": []\}.
  \item level1/level2/level3 must preferably use existing taxonomy nodes; if a new node is needed, set ontree=false and explain in reason.
  \item Do not merge multiple sources into a compound class, and do not repeat the caption text.
  \item Keep evidence within 12 characters and reason within 18 characters; output only the required fields.
  \item If level1 = Speech: level2/level3 must come from the Speech subtree, and a language field must be output, one of English / Mandarin Chinese / Japanese / Korean / Cantonese / Other. Default to English if no language is stated and no non-English line is quoted. Singing, tongue twisters, and animated characters also receive a language tag.
  \item Non-speech sounds must not have a language field.
\end{itemize}
\medskip

\{full taxonomy guideline\}
\medskip

Output schema: \{"sounds": [\{"sound\_id", "level1", "level2", "level3",
"ontree", "on\_screen", "evidence", "confidence", "reason"\}]\}
\medskip

\textbf{\rmfamily User:}\\
Classify the following caption. Output only the JSON object.\\
CAPTION: \{caption\}\\
JSON:
\end{promptbox}

\begin{promptbox}[title={Synchronization-oriented rewriting prompt}]
\label{box:rewrite}
\ttfamily\small\raggedright
You are an audio-visual caption rewriting assistant. You will be given a
caption describing a video clip's visuals and the sounds in it. Rewrite it
according to the rules below and output in English.
\medskip

\textbf{\rmfamily Concepts (understand first):}
\begin{itemize}
\setlength{\itemsep}{1pt}
\setlength{\parskip}{0pt}
\setlength{\topsep}{2pt}
  \item On-screen sound: the person/object making the sound is visible in the frame AND its visible action matches the sound (e.g., you see lips moving while hearing speech, or see a dog panting while hearing it).
  \item Off-screen sound: the source is NOT visible (e.g., ``an off-screen voice says...'', ``a narrator says...'', background music with no visible source).
  \item Audio-visual consistency: every sound traces to a visible source whose action matches the sound.
\end{itemize}
\medskip

\textbf{\rmfamily Goal:}
After rewriting, every sound described must be on-screen and audio-visually
consistent: wherever there is a sound, a visible person/object in the frame
must be its source, with its action matching the sound.
\medskip

\textbf{\rmfamily Steps (in order):}
\begin{itemize}
\setlength{\itemsep}{1pt}
\setlength{\parskip}{0pt}
\setlength{\topsep}{2pt}
  \item Step 1: List every sound mentioned (speech, dialogue, animal sounds, ambient sounds, music, etc.).
  \item Step 2: For each sound, decide if its source is visible (on-screen) or not (off-screen). On-screen with matching action: keep it, and make the ``visible source + action matches sound'' relationship explicit (e.g., ``his lips are moving'', ``the boom mic points at the speaker'', ``the dog pants as it is petted''). Off-screen: rewrite using the partner method (Step 3). Off-screen ambient/music with no possible visible source (e.g., background music): DELETE it entirely.
  \item Step 3: Partner method --- reassign the line/sound to a person already visible in the frame, so they become its on-screen source, and state that this person is producing it (lips moving / facing the other / action matching). If no suitable person exists, you may add a visible speaker, describing their appearance, position, and speaking action.
  \item Step 4: You may reword the caption and add reasonable details (appearance, position, action, sound direction) as long as all sounds end up on-screen and consistent. Do not invent sounds unrelated to the visuals.
\end{itemize}
\medskip

\textbf{\rmfamily Hard rules:}
\begin{itemize}
\setlength{\itemsep}{1pt}
\setlength{\parskip}{0pt}
\setlength{\topsep}{2pt}
  \item Output in English, but keep all spoken dialogue in its ORIGINAL language --- do not translate the words people say.
  \item The caption description must be in English even though the input caption is in Chinese; only the spoken dialogue stays in its original language.
  \item No off-screen/voice-over/invisible-source sounds may remain; delete unfixable ones (e.g., background music).
  \item Every remaining sound must map to a visible source with matching action.
  \item Keep the original descriptions of camera angle, movement, setting, and characters.
  \item Write a single, smooth, coherent paragraph. Output ONLY the rewritten caption --- no explanations or steps.
\end{itemize}
\medskip

\textbf{\rmfamily Example:}\\
Input (originally in Chinese): In a medium shot, a man in a baseball cap is
petting a large dog. Soft background music plays. An off-screen voice asks,
``Is Simon home?'', and a woman at the door answers, ``He's at work.''
\smallskip

Output: A medium shot shows a man in a baseball cap petting a large dog. On
screen, the man in the baseball cap, his lips moving, asks, ``Is Simon
home?''\,$^{\dagger}$ A woman in the doorway, facing him on camera, replies,
``He's at work.''\,$^{\dagger}$ A boom mic held by a crew member points toward
both speakers, and the dog pants audibly as it is petted, its panting matching
its visible movement.
\smallskip

(Note: the off-screen voice was reassigned to the visible man --- the partner
method --- the dialogue stays in its original language, and the background
music was deleted.)\\
{\rmfamily\footnotesize $^{\dagger}$\,Spoken in Chinese in the actual prompt; translated here for readability.}
\medskip

\textbf{\rmfamily Now begin:}\\
Rewrite this caption:\\
\{caption\}
\end{promptbox}
\section{SyncReward Details}
\label{app:impl}

\subsection{Detailed Architecture of Syncreward}
The encoder is Meta's Perception Encoder AV (\texttt{pe-av-large}), run in
bfloat16.
Each clip is divided into $S{=}18$ segments of $0.5$\,s with a stride of
$0.25$\,s (half overlap), so the interaction module receives
$2S{+}2 = 38$ tokens,
$[\texttt{REW};\, v_{1:S};\, \texttt{MOD};\, a_{1:S}]$.
Audio is resampled to $48$\,kHz. Segment embeddings are the mean-pooled
last hidden states of the encoder, with its projection heads retained.
Audio and video embeddings are linearly projected to width $768$ and processed
by a $3$-layer Pre-LN Transformer ($8$ heads, FFN ratio $4$, dropout $0.1$);
the reward is read out from the \texttt{REW} token by a two-layer MLP followed
by a sigmoid rescaled to $[0,2]$.

\subsection{Stage 1: temporal contrastive pretraining}
Stage~1 is trained on real videos from AudioSet and VGGSound. We score every
clip with ImageBind audio--visual similarity and discard clips scoring below
$0.2$, which removes most videos whose soundtrack is unrelated to the visual
content (e.g., background music or voice-over).
We optimize a symmetric InfoNCE loss over the flattened
$[B{\times}S,\, D]$ audio--visual similarity matrix, where co-occurring segments
are positives and all other segments on the same GPU serve as negatives; the
temperature is initialized at $0.07$ and learned.
We use AdamW with learning rate $1{\times}10^{-5}$, $1{,}000$ warmup steps,
gradient clipping at $1.0$, a per-GPU batch of $2$ clips, and an encoding
sub-batch of $6$.
We report the step-$5{,}500$ checkpoint.

\subsection{Stage 2: human alignment (reported model)}
We initialize the encoder from Stage~1 and keep it frozen; the interaction
module and reward head are randomly initialized and trained for $3$ epochs on
\texttt{SyncReward-Data-80K}.
The validation and test splits are disjoint from training.
Following Section~\ref{sec:method}, the objective is
\begin{equation}
\mathcal{L}
=
\mathrm{SmoothL1}(\hat r, r;\,\beta{=}0.1)
+
\lambda\,
\mathcal{L}_{\mathrm{rank}},
\qquad
\lambda{=}0.1,
\end{equation}
where $r\in[0,2]$ is the human synchronization score and $\mathcal{L}_{\mathrm{rank}}$ is a
softplus pairwise loss over in-batch pairs whose labels differ by at least
$0.75$.
We use AdamW with learning rate $1{\times}10^{-5}$, $100$ warmup steps, cosine
decay, and a per-GPU batch of $8$ on $2{\times}8$ H20 GPUs.
The released checkpoint is selected by validation
Spearman correlation.



\definecolor{hparamgray}{RGB}{238,238,238}

\begin{table}[t]
\centering
\caption{\textbf{SyncReward training hyperparameters.}
Settings for temporal pretraining (Stage~1) and human alignment
(Stage~2).}
\label{tab:sr-hparams}

\footnotesize
\setlength{\tabcolsep}{6pt}
\renewcommand{\arraystretch}{1.15}

\begin{tabularx}{\linewidth}{
    l
    >{\centering\arraybackslash}X
    >{\centering\arraybackslash}X
}
\toprule
\textbf{Setting} & \textbf{Stage 1} & \textbf{Stage 2} \\
\midrule

\rowcolor{hparamgray}
\multicolumn{3}{l}{\emph{Data and inputs}} \\
Training data
& AudioSet + VGGSound
& SyncReward-Data-80K \\
Filtering / labels
& ImageBind similarity $\geq 0.2$
& Human ratings: $0/1/2$ \\
Clip duration
& $4.75$\,s
& $4.75$\,s \\
Segmentation
& $18 \times 0.5$\,s
& $18 \times 0.5$\,s \\
Segment stride
& $0.25$\,s
& $0.25$\,s \\
Audio sample rate
& $48$\,kHz
& $48$\,kHz \\

\midrule
\rowcolor{hparamgray}
\multicolumn{3}{l}{\emph{Model}} \\
Encoder
& \texttt{pe-av-large}
& \texttt{pe-av-large} \\
Encoder updates
& Trainable
& Frozen \\
Trainable modules
& Encoders + projection heads
& Interaction + score head (23.5M) \\
Interaction Transformer
& --
& 3 layers, $d=768$, 8 heads \\

\midrule
\rowcolor{hparamgray}
\multicolumn{3}{l}{\emph{Training objective}} \\
Loss
& Symmetric InfoNCE
& Smooth $L_1$ + $0.1\,\mathcal{L}_{\mathrm{rank}}$ \\
Temperature
& Learned, initialized at $0.07$
& -- \\
Smooth $L_1$ parameter
& --
& $\beta=0.1$ \\
Ranking pair threshold
& --
& $|\Delta r| \geq 0.75$ \\

\midrule
\rowcolor{hparamgray}
\multicolumn{3}{l}{\emph{Optimization}} \\
Optimizer
& AdamW
& AdamW \\
Learning rate
& $1\times10^{-5}$
& $1\times10^{-5}$ \\
Cosine decay
& --
& Enabled \\
Warmup steps
& $1{,}000$
& $100$ \\
Gradient clipping
& $1.0$
& -- \\
Batch size per GPU
& $2$ (sub-batch: $6$)
& $8$ \\
Hardware
& $32\times$ H20
& $16\times$ H20 \\
Global batch size
& $64$
& $128$ \\
Training duration
& $5{,}500$ steps
& $3$ epochs \\
Checkpoint selection
& Final step
& Best validation Spearman $\rho$ \\

\bottomrule
\end{tabularx}
\end{table}


\section{Details of SyncReward-Bench}
\subsection{Benchmark Construction and Annotation}
\paragraph{Benchmark composition.}
SyncReward-Bench contains 1,000 generated clips from eight generators, with balanced representation across generators: 125 clips each. The benchmark covers four sound-event families: Speech (321), Onset (276), Instrument (169), and Ambient (234).

\paragraph{Human annotation.}
Each benchmark clip is independently assessed by three professional annotators following the same annotation guidelines. Annotators first assess audio--visual relatedness and then assign a synchronization score in $\{0,1,2\}$ to related clips. The evaluation target is the mean of the valid synchronization scores, without annotator-bias correction. 

\paragraph{Inter-annotator agreement.}
On the 1000 clips with three valid synchronization scores, Fleiss' $\kappa$ is $0.23$. Pairwise annotator comparisons yield Spearman correlations of $0.45$--$0.50$ and exact agreement rates of $52$--$55\%$. Opposite ratings ($0$ versus $2$) account for $1.3$--$2.3\%$ of paired judgments.

\paragraph{Evaluation metrics.}
We evaluate agreement with human ratings using Spearman's rank correlation and pairwise accuracy, accounting for each metric's score direction. Spearman's correlation compares metric scores with the aggregated human ratings. For pairwise accuracy, we consider all unordered pairs of distinct benchmark clips whose human ratings differ by at least $0.75$, regardless of generator or prompt. Each pair receives $1$ for a correct ordering, $0$ for an incorrect ordering, and $0.5$ for tied predictions. Pairwise accuracy is the average over all eligible pairs.

\subsection{Pairwise judge prompt for Gemini}
\begin{promptbox}[title={Pairwise judge prompt for Gemini-2.5-Pro}]
\ttfamily\small\raggedright
As a professional ``Text-to-Audible-Video'' quality inspector, your task is to
assess the quality of two videos ([[RESPONSE A]] and [[RESPONSE B]]) generated
from the same prompt ([[PROMPT]]). Each video carries its own generated audio
track. Rate the quality of the AI assistant's response(s) according to the
following criteria:
\medskip

\textbf{\rmfamily [Audio-Visual Synchronization]} Judge ONLY how well the audio
track of each video is synchronized with, and appropriate to, the visual
content of that same video:
\begin{itemize}
\setlength{\itemsep}{1pt}
\setlength{\parskip}{0pt}
\setlength{\topsep}{2pt}
  \item (1) Temporal alignment. Does each sound event start, peak and stop at the same moment as the on-screen action that would physically produce it (impacts, footsteps, speech and lip movement, collisions, mechanical clicks, door closes, water splashes, instrument strikes, etc.)? Penalize audio that leads or lags the corresponding visual event, drifts out of phase over the clip, or is smeared across the wrong frames.
  \item (2) Semantic correspondence. Do the sounds that are present match the actions, objects and materials that are visible? Penalize sounds with no visible cause, visibly sound-producing actions that stay silent, and sounds of the wrong type, wrong material or wrong number of events.
  \item (3) Realization of the sound events described in [[PROMPT]]. The prompt states which sounds should occur and at what point; prefer the video whose audio actually produces those events at the right moments.
\end{itemize}
\medskip

\textbf{\rmfamily Strict constraints:}
You MUST ignore all of the following; none of them may influence your
verdict: image quality, sharpness, resolution, bitrate, compression or codec
artifacts, colour, exposure, lighting, aesthetics, motion smoothness, visual
realism, artifacts or distortions in the picture, text rendering, audio
loudness or audio bitrate, and how faithfully the \emph{visual} content
follows the prompt. The two videos may differ greatly in visual quality while
being equally well synchronized; in that case they are equal on this
criterion. Judge audio-visual synchronization and nothing else.
\medskip

[[PROMPT]]: \{prompt\}
\medskip

\textbf{\rmfamily Output:}
Your score should reflect the quality of the AI assistant's response(s) with
respect to the specific criteria above, ignoring other aspects of the answer
(such as overall quality), and should agree with the score provided by a
reasonable human evaluator. The order of the responses is random, and you
must avoid letting the order bias your answer. Be as objective as possible in
your evaluation. Begin your evaluation by carefully analyzing the evaluation
criteria and the response. After providing your explanation, output your
final verdict by strictly following this format: ``[[A]]'' if response A is
better, ``[[B]]'' if response B is better, ``[[C]]'' means you cannot decide
which one is better (or they are equal). However, please try to avoid giving
a ``tie'' preference and be as decisive as possible.
\medskip

Return JSON only, with key ``reasoning'' (your analysis, citing the concrete
sound events and the timestamps at which they occur relative to the visuals)
and key ``verdict'' set to ``A'', ``B'', or ``tie'' (``tie'' corresponds to
``[[C]]'').
\end{promptbox}
\subsection{Human preference study instructions}
\begin{promptbox}[title={Human preference study instructions}]
\ttfamily\small\raggedright
The two videos are generated from the same text prompt, each with its own
generated audio track. Your task is to decide which video has \textbf{better
audio-visual synchronization}.
\medskip

\textbf{\rmfamily Criteria (judge synchronization only):}
\begin{itemize}
\setlength{\itemsep}{1pt}
\setlength{\parskip}{0pt}
\setlength{\topsep}{2pt}
  \item Temporal alignment: does each sound event start, peak and stop at the same moment as the on-screen action that physically produces it (impacts, footsteps, speech and lip movement, collisions, mechanical clicks, door closes, water splashes, instrument strikes, etc.)? Penalize sounds that lead or lag the visual event, drift out of phase, or are smeared across the wrong frames.
  \item Semantic correspondence: do the sounds match the visible actions, objects and materials? Penalize sounds with no visible source, visibly sound-producing actions that stay silent, and sounds of the wrong type, material, or number of events.
  \item Realization of the prompt: the prompt (shown above the videos) states which sounds should occur and when; prefer the video whose audio produces these events at the right moments.
\end{itemize}
\medskip

\textbf{\rmfamily Ignore:}
Image clarity, resolution, bitrate, compression artifacts, colour, exposure,
lighting, aesthetics, motion smoothness, visual artifacts and distortions,
text rendering, loudness, and how faithfully the \emph{visual} content follows
the prompt. Two videos may differ greatly in visual quality yet be equally
well synchronized; in that case they are equal under this criterion.
\medskip

\textbf{\rmfamily Procedure:}
\begin{itemize}
\setlength{\itemsep}{1pt}
\setlength{\parskip}{0pt}
\setlength{\topsep}{2pt}
  \item Wear headphones and make sure the volume is on; the task cannot be completed without sound.
  \item The two videos never play simultaneously: use ``Play in sequence'' to hear A and then B, and replay either video as often as needed.
  \item Choose ``A is better'', ``B is better'', or ``Tie''. Be as decisive as possible and select Tie only when the two cannot be distinguished.
  \item You may revisit and revise earlier answers; a revision overwrites the previous answer.
\end{itemize}
\end{promptbox}
\section{Generator Post-training}
\label{app:nft}

We post-train LTX-2.3 with DiffusionNFT~\citep{zheng2026diffusionnft}, an
online reinforcement learning method that optimizes a diffusion or flow
model directly on the forward (noising) process. It requires neither
likelihood estimation along the reverse trajectory nor a specific sampler.
This subsection summarizes the objective, the policy and sampler, the prompt
pool, and the reward arms; Table~\ref{tab:nft-hparams} lists the recipe.

\paragraph{DiffusionNFT objective.}
For a prompt $c$, the sampling policy generates a group of $k$ rollouts
$\{x_0^{(i)}\}_{i=1}^{k}$ that are scored by the reward model $R$.
Group-relative advantages are converted into optimality probabilities
\begin{equation}
A_i=\frac{R_i-\operatorname{mean}_{j\in\mathcal G(c)}R_j}{\operatorname{std}(R)},\qquad
r_i=\tfrac12+\tfrac12\operatorname{clip}\!\big(A_i/A_{\max},-1,1\big)\in[0,1],
\label{eq:nft_adv}
\end{equation}
where the mean is taken within the group of prompt $c$, the standard
deviation over all rewards in the sampling round, and $A_{\max}=5$.
Each rollout is re-noised along the forward process,
$x_t=(1-\sigma)x_0+\sigma\epsilon$ with $\epsilon\sim\mathcal N(0,I)$, and
the velocity prediction is converted into a clean-sample estimate
$\hat x_\theta=x_t-\sigma v_\theta(x_t,\sigma,c)$.
DiffusionNFT defines implicit positive and negative policies around a frozen
reference $\hat x_{\mathrm{old}}$:
\begin{equation}
x^{+}_\theta=(1-\beta)\,\hat x_{\mathrm{old}}+\beta\,\hat x_\theta,\qquad
x^{-}_\theta=(1+\beta)\,\hat x_{\mathrm{old}}-\beta\,\hat x_\theta ,
\label{eq:nft_policies}
\end{equation}
and minimizes
\begin{equation}
\mathcal L_{\mathrm{NFT}}(\theta)=
\mathbb E_{c,\,x_0,\,\sigma,\,\epsilon}\!\left[
r\,\frac{\|x^{+}_\theta-x_0\|_2^2}{w^{+}}
+(1-r)\,\frac{\|x^{-}_\theta-x_0\|_2^2}{w^{-}}
\right],
\label{eq:nft_loss}
\end{equation}
where $w^{\pm}$ is the per-sample mean absolute error of the corresponding
prediction, which acts as an adaptive weight.
High-reward rollouts pull the positive policy toward themselves.
Low-reward rollouts are fitted by the negative policy, which, by the mirror
symmetry of Eq.~\eqref{eq:nft_policies}, pushes $\hat x_\theta$ away from
them.
At the optimum, the learned policy moves from $v_{\mathrm{old}}$ along the
difference between the positive and negative data distributions. This
internalizes a guidance direction, so CFG is unnecessary at inference.
We use $\beta=1$, which gives $x^{+}_\theta=\hat x_\theta$ and
$x^{-}_\theta=2\hat x_{\mathrm{old}}-\hat x_\theta$, and weight the video and
audio latents equally.
The reference policy is updated once per sampling round by an exponential
moving average,
$\theta_{\mathrm{old}}\leftarrow\eta_t\,\theta_{\mathrm{old}}+(1-\eta_t)\,\theta$
with $\eta_t=\min(0.001\,t,\,0.5)$.
No KL penalty is used.

\paragraph{Policy and sampler.}
We start from the released distilled checkpoint and its $8$-step Euler schedule
$\sigma\in\{1, .994, .988, .981, .975, .909, .725, .422, 0\}$, with
classifier-free guidance and STG disabled for both video and audio.
The transformer, Gemma-3 text encoder, video VAE, audio VAE and vocoder are
frozen. A trainable LoRA ($r{=}32$, $\alpha{=}64$, Gaussian init, no dropout)
is attached to \texttt{to\_q}/\texttt{to\_k}/\texttt{to\_v}/\texttt{to\_out.0}
of all six attention types in each of the $48$ blocks: video and audio self-
and text-attention, and both audio--video cross-attentions ($214$M
parameters). Both the video and the audio stream are therefore updated.
Rollouts are $512{\times}768$, $121$ frames at $24$\,fps ($5.04$\,s) with
$48$\,kHz audio.
The loss in Eq.~\eqref{eq:nft_loss} is evaluated only at the four
highest-noise levels $\sigma\in\{1,.994,.988,.981\}$, with fresh noise for
every level.

\paragraph{Grouping and optimization.}
Each sampling round draws $2$ prompts with $k{=}32$ rollouts each
($16$ GPUs $\times$ $4$ rollouts, distinct seeds), and advantages are clipped
to $[-5,5]$.
Each round is consumed by $2$ optimizer steps (per-GPU micro-batch $2$, i.e.,
$32$ rollouts per step), and each rollout is used once.
$2{,}000$ steps therefore correspond to $64{,}000$ rollouts, or two passes
over the $1{,}000$ training prompts.
We use AdamW ($\beta_1{=}0.9$, $\beta_2{=}0.999$, weight decay $10^{-4}$,
$\epsilon{=}10^{-8}$) with a constant learning rate of $1{\times}10^{-5}$ and
no warmup, gradient clipping at $1.0$, bf16 autocast with fp32 master
weights, activation checkpointing, and FSDP full sharding on $2{\times}8$ H20
GPUs.
A run takes $100$--$162$ wall-clock hours ($180$--$291$\,s per step,
${\approx}1.6$--$2.6$k GPU-hours per arm).
We evaluate the final checkpoint (step $2{,}000$) of every arm without
checkpoint selection.

\paragraph{Prompt pool.}
Starting from $80k$ captions, a rule-based filter keeps only on-screen,
causally visible sound. It rejects background music, narration, off-screen or
ambient audio, stylized or animated content, applause, montage cuts and very
dark scenes; it also requires the source to be visible and caps quoted speech
so that it fits in $5$\,s.
Captions are assigned to four categories: multi-speaker dialogue, difficult
speech (singing, rapping, tongue-twisters), on-screen percussion, and sudden
impacts (glass, gunshots, hammer blows).
Using human $0$--$2$ synchronization ratings, we drop prompts on which
LTX-2.3 already scores above $1.5$. We prioritize prompts with evidence of
failure: LTX-2.3 rated at most $1.5$ or judged unrelated, or another
generator rated at most $1.0$. Such evidence exists for $99\%$ of the
training prompts.
With an equal per-category quota, $1{,}120$ prompts are selected and every
ninth prompt in the ranking is held out. This gives $1{,}000$ training
prompts ($255/252/246/247$ per category) and $120$ monitoring prompts
($25/28/34/33$).
Prompts from SyncReward-Bench and from the reward-model validation and
calibration splits are excluded.
The monitoring prompts are used only to track each arm's own reward during
training.

\begin{table}[t]
\centering
\caption{\textbf{DiffusionNFT post-training recipe.} }
\label{tab:nft-hparams}
\small
\begin{tabular}{ll}
\toprule
Generator & LTX-2.3 $22$B, distilled-1.1, $8$-step Euler, CFG/STG off \\
Resolution / length & $512{\times}768$, $121$ frames @ $24$\,fps, $48$\,kHz audio \\
Policy & LoRA $r{=}32$, $\alpha{=}64$, all attn.\ q/k/v/o (video+audio), $214$M \\
Group size $k$ / prompts per round & $32$ / $2$ ($64$ rollouts) \\
Micro-batch / rollouts per step & $2$ per GPU / $32$ \\
Advantage & group mean, batch std, clip $\pm5$, $A_{\max}{=}5$ \\
Trained $\sigma$ & $4$ highest of $8$ ($\sigma{\ge}0.98$) \\
NFT $\beta$ / reference EMA $\eta_t$ & $1.0$ / $\min(0.001t,0.5)$ per round \\
Optimizer & AdamW, lr $10^{-5}$ const., wd $10^{-4}$, clip $1.0$ \\
Precision / parallelism & bf16, FSDP full shard, activation ckpt. \\
Steps / rollouts & $2{,}000$ / $64{,}000$ ($2$ passes) \\
Train prompts & $1{,}000$ (four on-screen categories) \\
Monitoring prompts & $120$ (disjoint, same construction) \\
Hardware  & $16{\times}$H20  \\
\bottomrule
\end{tabular}
\end{table}

\end{document}